%% file: main_Paper.tex
\documentclass[runningheads]{llncs}

\usepackage{config/eccv}

\usepackage{config/eccvabbrv}

\usepackage{graphicx}
\usepackage{booktabs}

\usepackage[accsupp]{axessibility}  %

\usepackage[pagebackref,breaklinks,colorlinks,citecolor=eccvblue]{hyperref}

\input{config/preamble}

\input{config/definitions}

\begin{document}

\title{\vspace{-0.50 em}\textcolor{cvprblue}{LEXIS}: \textcolor{cvprblue}{L}at\textcolor{cvprblue}{E}nt pro\textcolor{cvprblue}{X}imal \textcolor{cvprblue}{I}nteraction \textcolor{cvprblue}{S}ignatures for 3D HOI from an Image\vspace{-0.50 em}}

\titlerunning{LEXIS}

\input{config/authors}

\maketitle

\input{fig/01_teaser}

\input{sec/00_Abstract}
\input{sec/01_Intro}

\input{sec/02_Related}
\input{sec/03_Method}
\input{sec/04_Experiments}
\input{sec/05_Conclusion}
\input{sec/06_Acknowledgments}

\pagebreak
\bibliographystyle{config/splncs04}
\bibliography{config/BIB}

\vfill
\clearpage
\begin{center}
  \vspace*{0.5em}
  {\LARGE\bfseries Supplementary Material}\\[0.6em]
  {\large\outTitleFULL}
\end{center}
\vspace{0.5em}
\input{sec/X_suppl}

\end{document}

%% file: config/preamble.tex
\usepackage{multirow}
\usepackage{pifont}
\usepackage{overpic}

\usepackage[utf8]{inputenc}
\usepackage[english,greek,main=english]{babel}

\usepackage{lipsum}
\usepackage{enumitem}
\usepackage{algorithmic}
\usepackage{algorithm}
\usepackage[normalem]{ulem} %
\usepackage{array}
\usepackage[capitalize]{cleveref}
\usepackage{wrapfig}
\usepackage{nicefrac}
\usepackage{fontawesome5} %

%% file: config/definitions.tex
\definecolor{cvprblue}{RGB}{0, 113, 188}

\newcommand{\orange}[1]{\textcolor{black}{#1}}
\newcommand{\magenta}[1]{{\textcolor{black}{#1}}}
\newcommand{\blue}[1]{{\textcolor{blue}{#1}}}

\renewcommand{\blue}[1]{#1}

\newcommand{\nameCOLOR}[1]{\textcolor{black}{#1}} %

\newcommand{\termCOLOR}[1]{\textcolor{black}{#1}} %

\newcommand{\lexisDictLONG}{{\textcolor{cvprblue}{L}at\textcolor{cvprblue}{E}nt pro\textcolor{cvprblue}{X}imal \textcolor{cvprblue}{I}nteraction \textcolor{cvprblue}{S}ignatures}\xspace}
\newcommand{\lexisDict}{\mbox{\nameCOLOR{LEXIS}}\xspace}
\newcommand{\lexisDictCode}{\mbox{\nameCOLOR{\lexisDict Code}}\xspace}
\newcommand{\lexisNet}{\mbox{\nameCOLOR{\lexisDict-Net}}\xspace}
\newcommand{\lexisFlow}{\mbox{\nameCOLOR{\lexisDict-Flow}}\xspace}
\newcommand{\lexisFlowRefinement}{\mbox{\nameCOLOR{\lexisDict-Flow$^*$}}\xspace}
\newcommand{\lexisFlowGaussian}{\mbox{\nameCOLOR{GaussFlow}}\xspace}
\newcommand{\flowPose}{\mbox{\nameCOLOR{FlowObject}}\xspace}
\newcommand{\interfield}[0]{\mbox{\nameCOLOR{InterField}}\xspace}
\newcommand{\interfields}[0]{\mbox{\nameCOLOR{InterFields}}\xspace}

\definecolor{darkorange}{HTML}{C96A00}
\newcommand{\metricbetter}[2]{#1~\makebox[4.0em][l]{\scriptsize\textcolor{darkorange}{(#2\%)}}}
\definecolor{ablgreen@iv}{HTML}{063A06}    %
\definecolor{ablgreen@iii}{HTML}{158015}   %
\definecolor{ablgreen@ii}{HTML}{1DC81D}  %
\definecolor{ablgreen@i}{HTML}{3EF03E}   %
\newcommand{\ablbetter}[3]{%
  \ifnum#1=1\colorlet{@ablc}{ablgreen@i}\fi
  \ifnum#1=2\colorlet{@ablc}{ablgreen@ii}\fi
  \ifnum#1=3\colorlet{@ablc}{ablgreen@iii}\fi
  \ifnum#1=4\colorlet{@ablc}{ablgreen@iv}\fi
  #2~\makebox[3.5em][l]{\scriptsize\textcolor{@ablc}{(#3\%)}}%
}

\newcommand{\outTitleFULL}{\vspace{-0.3 em}\textcolor{cvprblue}{LEXIS}: \textcolor{cvprblue}{L}at\textcolor{cvprblue}{E}nt pro\textcolor{cvprblue}{X}imal \textcolor{cvprblue}{I}nteraction \textcolor{cvprblue}{S}ignatures 
for 3D HOI from an Image
\vspace{-0.3 em}}

\newcommand{\numJoints}[0]{21}

\newcommand{\samtd}[0]{\mbox{SAM3D}\xspace}

\newcommand{\phosa}[0]{\mbox{\termCOLOR{PHOSA}}\xspace}

\newcommand{\chore}[0]{\mbox{\termCOLOR{CHORE}}\xspace}

\newcommand{\contho}[0]{\mbox{\termCOLOR{CONTHO}}\xspace}

\newcommand{\hoigaussian}[0]{\mbox{\termCOLOR{HOI-Gaussian}}\xspace}
\newcommand{\HOIGAUSSIAN}[0]{\hoigaussian}

\newcommand{\hdm}[0]{\mbox{\termCOLOR{HDM}}\xspace}

\newcommand{\tokenHMR}[0]{\mbox{\termCOLOR{TokenHMR}}\xspace}

\newcommand{\hoitg}[0]{\mbox{\termCOLOR{HOI-TG}}\xspace}

\newcommand{\sdedit}[0]{\mbox{\termCOLOR{SDEdit}}\xspace}
\newcommand{\SDEdit}[0]{\sdedit}

\newcommand{\chorus}[0]{\mbox{\termCOLOR{CHORUS}}\xspace}
\newcommand{\coma}[0]{\mbox{\termCOLOR{ComA}}\xspace}
\newcommand{\interactvlm}[0]{\mbox{\termCOLOR{InteractVLM}}\xspace}

\newcommand{\interactvlmSAMthreeD}[0]{\mbox{\nameCOLOR{InteractVLM++}}\xspace}
\newcommand{\pico}[0]{\mbox{\termCOLOR{PICO}}\xspace}
\newcommand{\picodb}[0]{\mbox{\termCOLOR{PICO-DB}}\xspace}

\newcommand{\camerahmr}[0]{\mbox{\termCOLOR{CameraHMR}}\xspace}
\newcommand{\CGHOI}[0]{\mbox{\termCOLOR{CG-HOI}}\xspace}
\newcommand{\ODE}[0]{\mbox{\termCOLOR{ODE}}\xspace}

\newcommand{\neuraldome}[0]{\mbox{\termCOLOR{NeuralDome}}\xspace}
\newcommand{\omomo}[0]{\mbox{\termCOLOR{OMOMO}}\xspace}
\newcommand{\imhd}[0]{\mbox{\termCOLOR{IMHD}}\xspace}
\newcommand{\interact}[0]{\mbox{\termCOLOR{InterAct}}\xspace}
\newcommand{\InterAct}[0]{\mbox{\interact}\xspace}

\newcommand{\mocap}[0]{\mbox{\termCOLOR{MoCap}}\xspace}

\newcommand{\pose}[0]{\boldsymbol{\theta}\xspace}
\newcommand{\shape}[0]{\boldsymbol{\beta}\xspace}

\newcommand{\bc}{\mathbf{c}}

\newcommand{\bff}{\mathbf{f}} %

\newcommand{\bI}{\mathbf{I}}

\newcommand{\bp}{\mathbf{p}}\newcommand{\bP}{\mathbf{P}}

\newcommand{\bu}{\mathbf{u}}
\newcommand{\bv}{\mathbf{v}}\newcommand{\bV}{\mathbf{V}}

\newcommand{\bx}{\mathbf{x}}

\newcommand{\nR}{\mathbb{R}}

\newcommand{\cB}{\mathcal{B}}
\newcommand{\cC}{\mathcal{C}}
\newcommand{\cD}{\mathcal{D}}
\newcommand{\cE}{\mathcal{E}}

\newcommand{\cI}{\mathcal{I}}

\newcommand{\cL}{\mathcal{L}}
\newcommand{\cM}{\mathcal{M}}
\newcommand{\cN}{\mathcal{N}}
\newcommand{\cO}{\mathcal{O}}

\newcommand{\cS}{\mathcal{S}}
\newcommand{\cT}{\mathcal{T}}

\newcommand{\cZ}{\mathcal{Z}}

\newcommand{\bzero}{\mathbf{0}}

\DeclareOldFontCommand{\bf}{\normalfont\bfseries}{\mathbf}

\newcommand{\bepsilon}{\boldsymbol{\epsilon}}

\newcommand{\VQVAE}{\mbox{VQ-VAE}\xspace}

\newcommand{\IMU}{\mbox{IMU}\xspace}
\newcommand{\RGB}{\mbox{RGB}\xspace}
\newcommand{\RGBD}{\mbox{RGB-D}\xspace}
\newcommand{\GT}{{GT}\xspace}
\newcommand{\pGT}{\mbox{pseudo-GT}\xspace}

\DeclareSymbolFont{matha}{OML}{txmi}{m}{it}%
\DeclareMathSymbol{\varv}{\mathord}{matha}{118}

\newcommand{\NA}{\mbox{\textcolor{gray}{N/A}}\xspace}

\renewcommand{\ie}{\mbox{i.e.}\xspace}
\renewcommand{\eg}{\mbox{e.g.}\xspace}
\renewcommand{\wrt}{\mbox{w.r.t.}\xspace}

\newcommand{\qheading}[1]{\noindent\textbf{#1:}}
\newcommand{\zheading}[1]{\textbf{#1:}}

\newcommand{\qheadingFIGTAB}[1]{\textbf{#1.}}

\newcommand{\smpl}{\mbox{\termCOLOR{SMPL}}\xspace}

\newcommand{\HOI}{\mbox{\termCOLOR{HOI}}\xspace}
\newcommand{\smplx}{\mbox{SMPL-X}\xspace}

\newcommand{\smplh}{\mbox{SMPL+H}\xspace}
\newcommand{\smplH}{\smplh}
\newcommand{\SMPLH}{\smplh}

\newcommand{\sota}[0]{\mbox{SotA}\xspace}

\newcommand{\inthewild}{{in-the-wild}\xspace}
\newcommand{\Inthewild}{{In-the-wild}\xspace}

\newcommand{\itw}[0]{in-the-wild\xspace}
\newcommand{\video}{\textcolor{magenta}{{{website video}}}\xspace}

\newcommand{\procigen}{\mbox{ProciGen}\xspace}

\newcommand{\opentdhoi}{\mbox{Open3DHOI}\xspace}

\newcommand{\behave}{\mbox{BEHAVE}\xspace}
\newcommand{\BEHAVE}{\behave}

\newcommand{\intercap}{\mbox{InterCap}\xspace}

\newcommand{\myhat}[1]{\hat{#1}}
\newcommand{\triplaneHum}{\cT^\humSuper}
\newcommand{\triplaneObj}{\cT^\objSuper}
\newcommand{\triplaneRes}{T_r}
\newcommand{\triplaneChan}{T_c}
\newcommand{\triplaneDim}{3 \times \triplaneRes \times \triplaneRes \times \triplaneChan}

\newcommand{\lexisTokens}{\cZ}
\newcommand{\lexisTokensPRED}{\myhat{\cZ}}
\newcommand{\lexisDim}{21 \times D}

\newcommand{\humSuper}{{b}}
\newcommand{\objSuper}{{o}}

\newcommand{\humR}{R^\humSuper}
\newcommand{\humt}{t^\humSuper}
\newcommand{\objR}{R^\objSuper}
\newcommand{\objt}{t^\objSuper}
\newcommand{\img}{\cI}                      %
\newcommand{\objPC}{\bP^\objSuper}                  %
\newcommand{\humState}{\cB}                 %
\newcommand{\objState}{\cO}                 %
\newcommand{\hoiState}{\bx}                 %
\newcommand{\codebook}{\cC}                 %
\newcommand{\centry}{\bc}                   %
\newcommand{\surfpts}{\cS}                  %
\newcommand{\querypt}{\bp}                  %
\newcommand{\noise}{\bepsilon}              %
\newcommand{\objverts}{\bV^\objSuper}                 %
\newcommand{\lexisDecoder}{\mbox{\nameCOLOR{$D_\psi$}}\xspace}
\newcommand{\estV}{\widehat{\bV}}
\newcommand{\estVertex}{\hat{\bv}}
\newcommand{\gtV}{\bV}
\newcommand{\gtVertex}{\bv}

\newcommand{\objSurface}{\cO}
\newcommand{\objSurfacePoint}{p^\objSuper}
\newcommand{\humSurfacePoint}{p^\humSuper}
\newcommand{\estMask}{\hat{M}}
\newcommand{\gtMask}{M}
\newcommand{\indicatorFunc}{\mathbb{I}}
\newcommand{\medianHumDepth}{\bar{z}^\humSuper}
\newcommand{\humDepth}{z^\humSuper}
\newcommand{\medianObjDepth}{\bar{z}^\objSuper}

\newcommand{\humVerts}{\bV^\humSuper}
\newcommand{\interFieldPRED}{\nameCOLOR{\myhat{\text{IF}}}}          %
\newcommand{\interFieldLoss}{\cL_{IF}}          %
\newcommand{\interFieldGuideLoss}{\cL_{\text{pose+IF}}} %
\newcommand{\interFieldLossWeight}{\lambda_\text{IF}}          %
\newcommand{\interFieldGT}{\text{IF}} %
\newcommand{\zhat}{\myhat{\cZ}}              %
\newcommand{\objfeat}{\bff^\objSuper}      %
\newcommand{\velfield}{v_{\gamma}}            %
\newcommand{\vtarget}{\bu_{\text{target}}}  %
\newcommand{\objmask}{M^\objSuper}                    %
\newcommand{\proj}{\pi}                     %
\newcommand{\guidestep}{\eta}               %
\newcommand{\maskWeight}{\lambda_{\text{mask}}} %

\newcommand{\posePred}{\myhat{\pose}}          %
\newcommand{\poseGt}{\pose}              %

%% file: config/authors.tex
\authorrunning{D.~Anti\'{c} et al.}

\author{
    Dimitrije Anti\'{c}$^1$                     \quad 
    Alvaro Budria\textsuperscript{*}$^1$        \quad 
    George Paschalidis\textsuperscript{*}$^1$   \\    
    Sai Kumar Dwivedi$^2$                       \quad 
    Dimitrios Tzionas$^{1,3}$
}
\institute{
    {
        \scriptsize
        {* Equal contribution}                                                  \\
        $^1$University of Amsterdam, The Netherlands                            \\
        $^2$Max Planck Institute for Intelligent Systems, T{\"u}bingen, Germany \\
        $^3$Aristotle University of Thessaloniki, Greece                        \\
    }
    \vspace{+0.4 em}
}

%% file: fig/01_teaser.tex
\begin{figure}
    \vspace{-2.5 em}
    \centering      %
    \includegraphics[trim=000mm 000mm 000mm 000mm, clip=true, width=0.99 \textwidth]{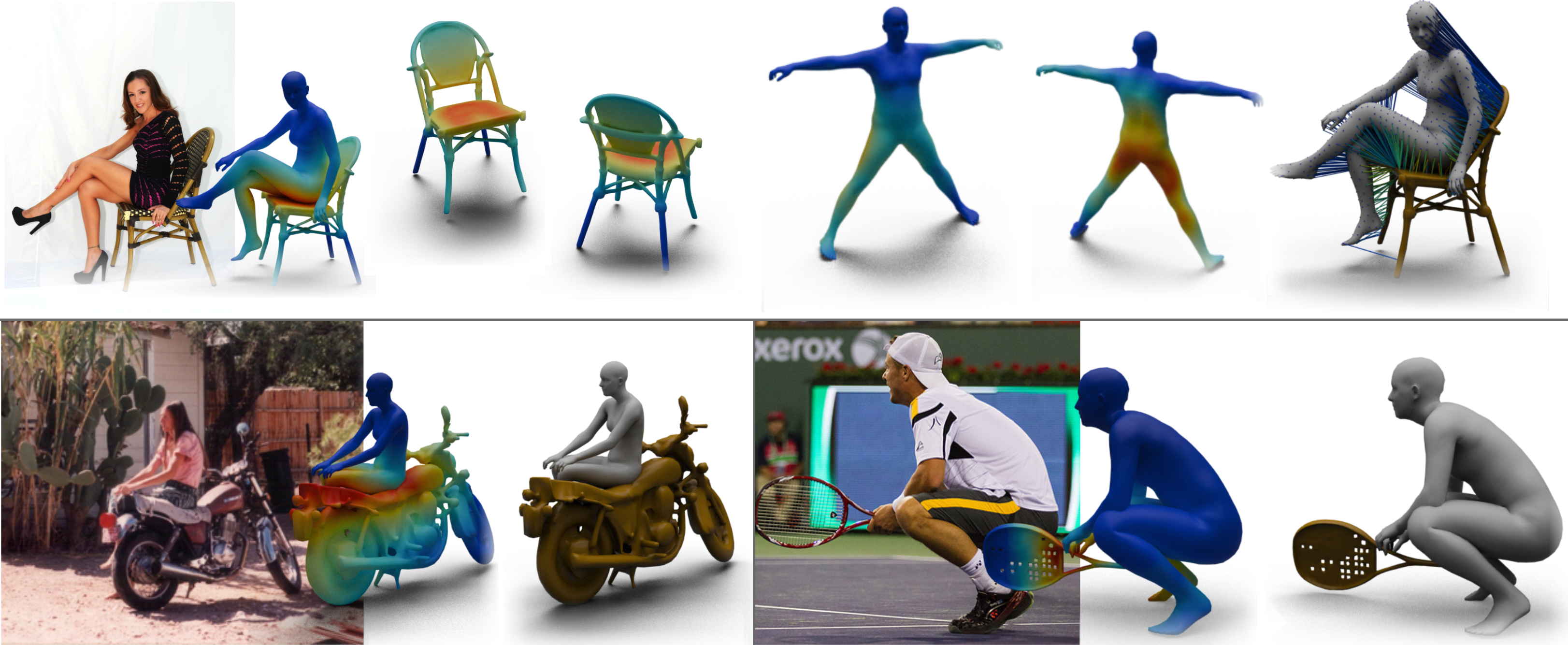}
    \vspace{-0.9 em}
    \caption{
        We present \lexisFlow, a framework for 3D Human-Object Interaction (\HOI) reconstruction from single images. 
        We go beyond sparse, binary contact by learning \lexisDict, a latent manifold of dense, continuous interaction signatures. 
        \lexisDict-guided sampling enables recovering physically-plausible \HOI without post-hoc optimization.
    }
    \label{fig:teaser}
    \vspace{-2.3 em}
\end{figure}

%% file: sec/00_Abstract.tex
\begin{abstract}
Reconstructing 3D Human–Object Interaction from an \RGB image is essential for perceptive systems. 
Yet, this remains challenging as it requires capturing the subtle physical coupling between the body and objects. 
While current methods rely on \emph{sparse, binary contact} cues, these fail to model the continuous proximity and dense spatial relationships that characterize natural interactions.
We address this limitation via \interfields, a representation that encodes \emph{dense, continuous proximity} across the entire body and object surfaces. 
However, inferring these fields from single images is inherently ill-posed. To tackle this, our intuition is that interaction patterns are characteristically structured by the action and object geometry. 
We capture this structure in \lexisDict, a novel discrete manifold of interaction signatures learned via a \VQVAE.
We then develop \lexisFlow, a \blue{diffusion} framework that leverages \lexisDict signatures to estimate human and object meshes alongside their \interfields. 
Notably, these \interfields~help in a guided refinement that ensures physically-plausible, proximity-aware reconstructions without requiring post-hoc optimization. 
Evaluation on \opentdhoi and \BEHAVE shows that \lexisFlow significantly outperforms existing \sota baselines in reconstruction, contact, and proximity quality. 
Our approach not only improves generalization but also yields reconstructions perceived as more realistic, moving us closer to holistic 3D scene understanding.
Code \& models will be public at \url{https://anticdimi.github.io/lexis}.  
\enlargethispage{2.5 em}
\end{abstract}

\pagebreak

%% file: sec/01_Intro.tex
\section{Introduction}
\label{sec:intro-refined}

\input{fig/02_motivation}

Our actions are inherently defined and physically constrained bythe objects around us. 
Accurately recovering 3D human–object interaction (\HOI) 
from a single \RGB image is essential for virtual and robotic assistants, mixed reality, animation, and analyzing interactions from internet-scale images. 
Technically, this task involves estimating not only the 3D pose and shape of the human and the object, but also their relative 3D \emph{spatial configuration} and subtle \emph{physical coupling}.

This is challenging due to depth/scale ambiguities, self-occlusions, and the mutual occlusions between bodies and objects in RGB images. 
These ambiguities pose significant hurdles for computational methods. 
Yet, humans navigate these challenges effortlessly via prior experience. 
Capturing this human-like capability requires two components: a \emph{prior model} that captures the interaction ``geometry,'' and a mechanism to \emph{integrate} this prior into the \emph{reconstruction process}.

Despite their importance, both components involve open problems because we currently lack a representation that can bridge the gap between sparse 2D image cues and dense 3D physical interaction. 
To tackle this, past work exploits \emph{contact} \cite{dwivedi_interactvlm_2025, GRAB:2020, yang2024lemon, contactgen, shimada2022hulc, brahmbhatt2020contactPose}; 
body and object surface points are classified as in contact or not. 
Contact points act as anchors that align body–object geometry, accounting for the occlusions and depth ambiguities of images. 

However, contact has \blue{fundamental} limitations. 
First, it is only a \emph{sparse} signal; non-contacting points are ignored. 
Second, it is only \emph{binary}; it encodes areas of zero mesh distance, ignoring non-zero ones. 
Think of a person working out (see \cref{fig:intro_toy}); 
contact between the body and objects stays fixed across time, failing to encode the interaction’s progression. 

To tackle the above limitations, we need a richer representation. 
Our key observation is that \emph{distances} between body points and object points change as the interaction progresses. 
This yields a \emph{dense}, \emph{continuous} signal of surface-to-surface proximity, namely a 3D \emph{Interaction Field}, or \interfield{} for short (see \cref{fig:teaser}). 
Note that contact is only a sparse subset (zero level set) of the broader interaction field; that is, the \interfield does not ``replace'' contact but instead ``\emph{extends}'' it. 
Intuitively, the \interfield representation captures geometry- and proximity-aware \emph{interaction signatures} that help 3D reconstruction. 

Yet, this potential remains underexplored. 
Existing work estimates hand-only \interfields \cite{fan2023arctic} from an image, without using them in any task. 
Other work exploits hand-only \interfields \cite{goal, manipnet} to synthesize grasps, but only ``implicitly'' for training losses. 
No existing method uses \interfields to guide \emph{neural in-model refinement} at \emph{inference} time.
We fill this gap by inferring both body and object \interfields from an image, and exploiting these for 3D reconstruction. 

\pagebreak  

However, inferring 3D fields, \ie, distances for all points across two 3D surfaces \blue{from just a 2D image} is \blue{inherently ill-posed}.  
Moreover, \blue{the space of possible interactions is vast and highly non-convex}.
To tackle \blue{these challenges}, 
we empirically observe that \interfields contain interaction- and object-specific patterns, or ``\emph{interaction signatures}.'' 
Thus, we learn a novel \blue{\emph{lexicon}} of these patterns called \lexisDict (\lexisDictLONG), \blue{namely a discrete \emph{manifold} of interaction signatures, encoding \emph{prior knowledge} about these.}
Specifically, to learn \lexisDict we train a \VQVAE \cite{dwivedi_cvpr2024_tokenhmr, fiche2024vq} that takes as input 3D human and object geometry in interaction, encodes this into a latent code, and decodes this into body and object \interfields. 
As training data we use \mocap-based, \blue{synthetically-augmented} 3D geometry of interactions. 

We exploit \lexisDict to estimate 3D interactions from a single \RGB image. 
Existing methods mostly follow a two‑stage pipeline: they first estimate coarse human and object shape/pose (often missing contacts), and then refine them in post-hoc optimization with contact-based losses to improve interaction plausibility. 
This is suboptimal for two reasons;  
first, contact provides only \emph{sparse, binary} cues; 
second, these are considered only ``\emph{too late},'' \ie, post-hoc. 
Instead, we leverage \emph{dense, continuous} \interfields and consider them ``\emph{early on}'' in our model. 
More specifically, we develop \lexisFlow, a Flow-Matching 
model that jointly estimates \blue{posed} 3D human and object meshes, along with their \interfields. 
Crucially, the predicted \interfields~help in a guided  refinement of meshes, eliminating the need for post-hoc optimization in a separate stage. 

We evaluate primarily on the \opentdhoi~\cite{open3dhoi} dataset (unless otherwise specified) 
due to its focus on \emph{in-the-wild} images and \emph{generalizability}. 
Evaluation shows that dense, continuous \interfields are \orange{more helpful} than sparse, binary contacts for formulating 3D reconstruction constraints; we test this both in \lexisFlow's \magenta{guided refinement} and in a fitting framework~\cite{dwivedi_interactvlm_2025}. 
Benchmarking shows that \lexisFlow \orange{clearly outperforms} \sota methods on 3D reconstruction in the wild and its estimates are perceived as significantly more realistic. 
In-distribution evaluation on the \behave~\cite{bhatnagar22behave} dataset \blue{echoes the above}. 
Ablation studies support our key design choices. 
Last, evaluation shows that 
\lexisFlow can be initialized with 3D estimates provided by \sota tools, and \magenta{\lexisFlow's guided refinement} \orange{clearly improves} these. 

In summary, here we make the following contributions:
\begin{enumerate}[itemsep=6pt,topsep=6pt]%
\item We go beyond sparse, binary contact by inferring \emph{dense, continuous} \interfields that encode \emph{proximity} cues across entire body and object surfaces. 
\item We learn the novel %
\lexisDict \emph{dictionary} that encodes action- and object-aware \emph{\blue{latent} interaction signatures}, which are learned from 3D interaction datasets. 
\item We develop \lexisFlow, a model that 
exploits \lexisDict to estimate a 3D human, object, and their \interfields, from an image. 
Notably, \lexisDict-based \interfields guide the Flow-Matching generative sampling to refine estimates and 
improve the 3D \HOI plausibility, without post-hoc optimization. 
\end{enumerate}
Code and models will be available at \url{https://anticdimi.github.io/lexis}.

%% file: fig/02_motivation.tex
\begin{wrapfigure}{r}{0.47 \textwidth}
    \centering
    \vspace{-2.40 em}    %
    \includegraphics[trim={00mm 36mm 00mm 00mm}, clip,width=0.47 \columnwidth]{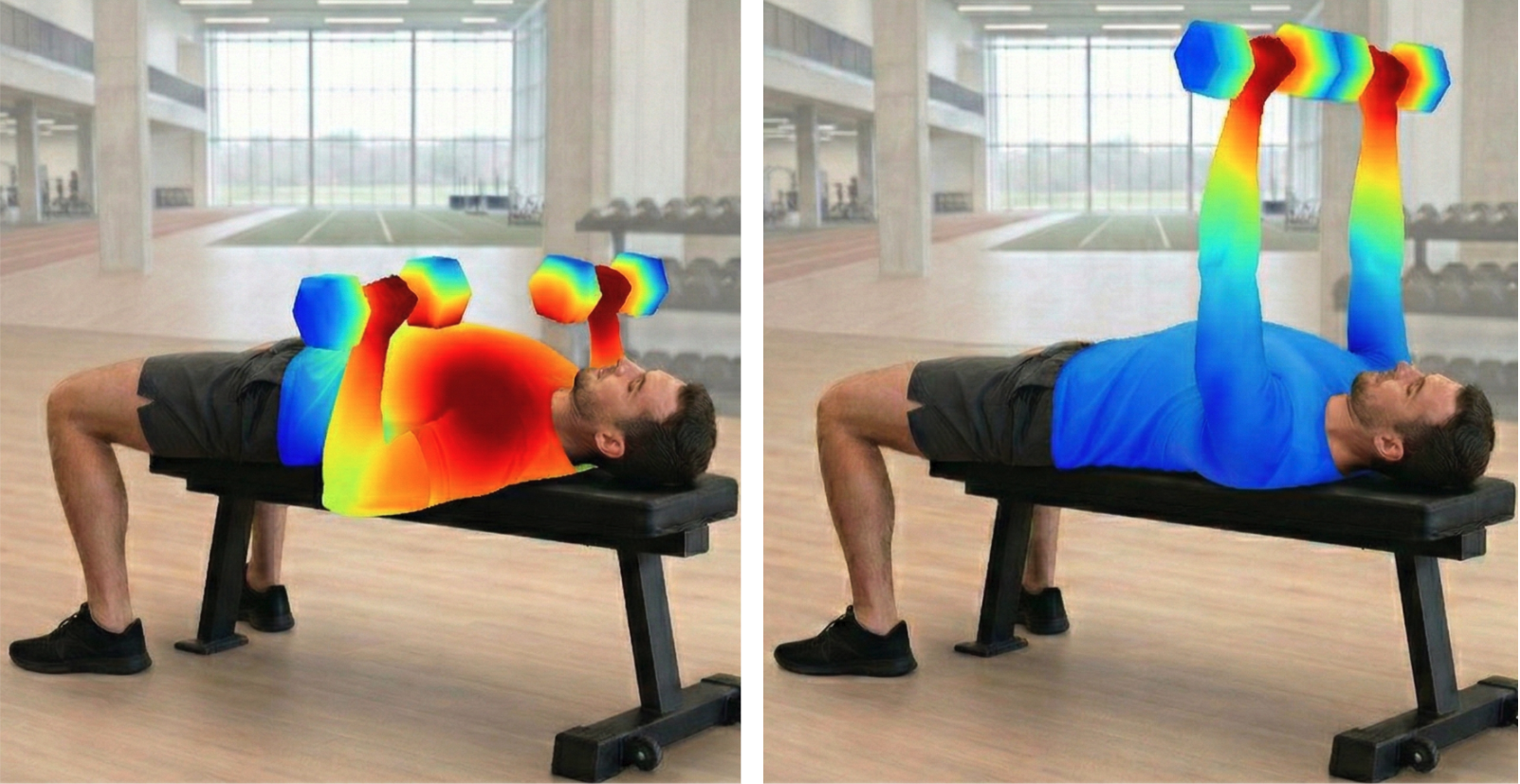}~
    \vspace{-0.80 em}
    \caption{
        \qheadingFIGTAB{Toy example}
        Contact between the body and bench, and between hands and dumbbells, remains fixed. 
        Instead, \emph{distances} from each body point to the objects, 
        v.v., form a rich, geometry- and proximity-aware \emph{interaction signature}. 
    }
    \vspace{-1.5 em}
    \label{fig:intro_toy}
\end{wrapfigure}

%% file: sec/02_Related.tex
\section{Related Work}
\label{sec:related_work}

\subsection{HOI Representations}

\hspace{\parindent}
\zheading{Binary Contact} Contact is a binary cue (points are in contact or not) that is used to taxonomize grasps \cite{dillmann2005grasp, Feix_GRASP_2016, kamakura1980grasp}, to synthesize hand \cite{GRAB:2020, jiang2021handobjectcc, contactgen, bimart} or body \cite{petrov2025tridi, hassan2021posa, kim2024coma, diller2024cghoi} interaction, and reconstruct hand \cite{grady2021contactopt, hampali2020honnotate, Yang_2021_CPF} or body \cite{zhang2020phosa, xie2022chore, nam2024contho} interaction from pixels. 
For the latter, existing work: 
(i)   detects contact pixels~\cite{chen2023hot}, 
(ii)  thresholds distances to compute 3D contact       \cite{huang2022rich, GRAB:2020}, 
(iii) exploits manual 3D contact labels \cite{cseke_tripathi_2025_pico, zhang2020phosa},  
(iv)  infers 3D contact \cite{dwivedi_interactvlm_2025, tripathi2023deco, nam2024contho} or 
(v)   pressure \cite{tripathi2023ipman, grady2022pressurevision}. 

\zheading{Spatial Probability}
Recent work uses text-to-image diffusion 
to generate 3D training data \cite{han2023chorus, kim2024coma}. 
\chorus~\cite{han2023chorus} learns a 3D object occupancy field \wrt the body. 
\coma~\cite{kim2024coma} extends this by learning an object-to-human 3D distance field with orientation cues. 
These works learn separate models per object/action class, and need heavy sampling/filtering, but their zero-shot nature is promising. 

\zheading{Interaction Fields (\interfields)}
Going beyond sparse, binary contact, \interfields encode dense, continuous surface-to-surface distances. 
\chore \cite{xie2022chore} thresholds learned Distance Fields (DF) to extract binary contacts. 
Other work uses DFs to synthesize body \cite{diller2024cghoi} or hand interaction \cite{goal, bimart, manipnet, taheri_3dv2024_grip, Xue2025rog} or to reconstruct grasps \cite{fan2023arctic, Yang_2021_CPF, HOISDF:CVPR:2024}, but mostly implicitly in features or losses. 
\lexisFlow differs from these in two ways: 
\textbf{(1) Usage:} 
\lexisFlow is the first method to use full-body \interfields explicitly in \emph{inference}. 
\textbf{(2) Task:} 
\lexisFlow does the above for \emph{reconstruction} (with \interfield-guided refinement), while \CGHOI \cite{diller2024cghoi} for synthesis.
Uniquely, \lexisFlow exploits a novel \emph{dictionary} of \interfields. 

\subsection{3D \HOI Reconstruction from a Single Image}

\hspace{\parindent}
\zheading{Human\blue{-only} Reconstruction} Optimization methods fit a parametric 3D body model~\cite{SMPL:2015, xu2020ghum, Joo2018_adam} to image cues \mbox{\cite{bogo2016simplify, Joo2018_adam}}. 
Regression-based methods estimate parameters of such models \cite{hmrKanazawa17, poco, camerahmr, VIBE:CVPR:2020, SAM3DBody::2025} or a non-parametric mesh \cite{lin2021graphformer, corona2022lvd, nlf2024sarandi}. 

\zheading{Object\blue{-only} Reconstruction}
Pose is estimated via regression \cite{meshrcnn,rw_pose1} or by fitting a template \cite{rw_pose3, rw_pose2, rw_pose4} or implicit SDF \cite{antic2025sdfit} with optimization. 
Shape is encoded as voxels~\cite{rw_voxel2, rw_voxel1}, point clouds~\cite{rw_pc1, rw_pc1}, superquadrics \cite{paschalidou2019superquadrics}, meshes~\cite{meshrcnn, pixel2mesh}, or neural fields \cite{deepsdf, dit, cheng2023sdfusion}. 
Recent work uses diffusion \cite{one-2-3-45, magic123, liu2023zero1to3} or database retrieval \cite{liu2023openshape, huand2023shapeclipper}. 
Recently \mbox{SAM3D} \cite{SAM3D::2025} estimates robust object shape, but its pose estimation lacks interaction awareness \wrt humans; 
we \emph{initialize} \lexisFlow with \mbox{SAM3D} object estimates, and \emph{refine pose} by adding interaction awareness. 

\zheading{\HOI Datasets}
Images paired with 3D \GT are scarce. 
Most datasets are captured in constrained in-lab \cite{fan2023arctic, zhang2023neuraldome}, indoor \cite{bhatnagar22behave, huang2024intercap, PROX:2019, savva2016pigraphs} or outdoor \cite{huang2022rich} settings. 
Sometimes these~\cite{bhatnagar22behave, huang2024intercap} are synthetically augmented~\cite{xie2024hdm_procigen}. 
To recover better 3D \pGT, the \behave \cite{bhatnagar22behave} and \intercap \cite{huang2024intercap} datasets are built using multi-view \RGBD cameras, while other datasets are built with \blue{multi-view} \RGB cameras \cite{huang2022rich, zhang2023neuraldome, zhao2024imhoi} along with \IMU sensors \cite{zhao2024imhoi}. 
Recently 3D contact labels are crowd-sourced \cite{tripathi2023deco, cseke_tripathi_2025_pico, open3dhoi, yang2024lemon} to help 3D reconstruction for in-the-wild images. 

\zheading{Template-free \HOI methods}
\hdm~\cite{xie2024hdm_procigen} \blue{infers} human and object point clouds via a hierarchical diffusion framework; this first predicts rough 
a point cloud for each, and then refines these separately via cross-attentive diffusion. 

\zheading{Template-based \HOI methods}
Object shape is often assumed known. 
Optimization \cite{zhang2020phosa,xie2022chore,open3dhoi} fits 3D human/object poses to images while satisfying contact constraints. 
\phosa~\cite{zhang2020phosa} assumes known 3D contacts. 
\HOIGAUSSIAN~\cite{open3dhoi} derives contacts from Gaussian splat opacities. 
\chore~\cite{xie2022chore} infers distance fields around the human/object and extracts from these binary contacts. 
Instead of optimizing, \contho~\cite{nam2024contho} directly \blue{infers} a human mesh and object orientation.

Most methods work for in-distribution images, captured in constrained environments with pre-scanned objects, so they struggle generalizing.
To tackle this, recent work annotates \emph{in-the-wild} images with 3D contact labels (at a part- \cite{open3dhoi} or vertex-level \cite{cseke_tripathi_2025_pico, tripathi2023deco} granularity), used to constrain optimization \cite{open3dhoi, cseke_tripathi_2025_pico}. 
For \blue{unlabeled} images, \pico \cite{cseke_tripathi_2025_pico} first infers (noisy) body-only 3D contact from pixels \cite{tripathi2023deco}, and uses this as query to \blue{retrieve} clean contact labels from its \picodb database. 
\interactvlm \cite{dwivedi_interactvlm_2025} learns contact from scarce 3D labels via vision-language models.  
\blue{\pico and \interactvlm retrieve shape} from a large 3D object database \cite{objaverse} via a joint image-geometry latent space \cite{liu2023openshape}, but databases have finite sizes.  
Instead, we estimate shape via the \mbox{SAM3D} \cite{SAM3D::2025} foundational model. 
Most work \cite{zhang2020phosa, xie2022chore, nam2024contho, cseke_tripathi_2025_pico, dwivedi_interactvlm_2025,wanghoi-tg} follows a \mbox{\emph{two-stage}} approach; 
it first estimates rough meshes, and then refines these via \emph{sparse, binary} contact in \emph{post-hoc} optimization \cite{zhang2020phosa, xie2022chore} or in a transformer \cite{nam2024contho}. 
Instead, we follow a \mbox{\emph{one-stage}} approach; 
we use diffusion to \emph{jointly} infer a 3D body, object, and respective \emph{dense, continuous} \interfields, which guide sampling to refine estimates. 

%% file: sec/03_Method.tex
\section{Methodology}
\label{sec:method}

\newcommand{\footnoteBodyModel}{\blue{Our method works directly for \smplH \cite{MANO:SIGGRAPHASIA:2017} \& \smpl \cite{SMPL:2015}.}} 

\hspace{\parindent}
\qheading{\blue{Overview}}
Given an \RGB image depicting a human--object interaction (\HOI), $\img \in \nR^{H \times W \times 3}$, and a 3D object shape estimation by an off-the-shelf model, the goal is to estimate the \HOI in 3D, namely to estimate a 3D human mesh (\smplh\footnote{\footnoteBodyModel} body \cite{MANO:SIGGRAPHASIA:2017}) and object mesh, shaped and posed such that they interact realistically and match image cues. 
Technically, this estimates two states: 
\noindent
    \textbf{(1) Object state $\objState = \{\objR,\, \objt\}$} in a \textbf{human-root-relative frame}, where
    $\objR \in \nR^{6}$         is the global rotation (in 6D form \cite{zhou2019rotations}), and 
    $\objt \in \nR^{3}$         the global translation;

\noindent
    \textbf{(2) Body state $\humState = \{ \humR,\, \humt,\ \shape,\ \lexisTokens \}$} in the \textbf{camera frame}, 
    where 
    \mbox{$\humR \in \nR^{6}$}                     is the global rotation (in 6D form \cite{zhou2019rotations}), %
    \mbox{$\humt \in \nR^{3}$}                     is the global translation, 
    \mbox{$\shape \in \nR^{10}$}                   are \SMPLH shape parameters, and 
    \mbox{$\lexisTokens \in \nR^{\lexisDim}$}      are 
    tokens of a \emph{novel} lexicon, called \lexisDict (see \cref{subsec:lexisnet}), 
    encoding both pose and ``\emph{interaction signatures}.'' 

\newcommand{\footnoteInterfieldDIMs}{\interfields are \textbf{high-dimensional}; 
to store a distance per \smplH vertex: $\nR^{6890}$.}

\zheading{\blue{Representation (\interfield)}}
To encode interaction relationships, most work uses \emph{sparse, binary contact}; points are either in contact or not. 
Recent work uses \interfields, $\interFieldGT(\querypt) \in \nR_{\geq 0}$, which, for every\footnote{\footnoteInterfieldDIMs} 3D body/object point, $\querypt$, encodes the distance to the nearest point on the interacting counterpart. 
This is a \emph{dense}, \emph{continuous} signal, that models both contact and \emph{proximity}. 
Our approach builds on this richer representation. 
However, estimating 3D \interfields from a 2D image is a highly ill-posed, high-dimensional problem. 
We tackle this below. 

\input{fig/03_lexisnet}

\zheading{Approach}
We develop a framework that learns compact ``\emph{interaction signatures}'' and \emph{explicitly} exploits these for 3D \HOI reconstruction from an image:

\noindent
\qheading{Interaction Signatures (\cref{subsec:lexisnet})}
We employ a \VQVAE and learn \textbf{\lexisDict}, a novel lexicon of \emph{compact interaction signatures} that encode interaction- and object-specific patterns observed in \interfields. 
Note that \lexisDict codes can be decoded into full 3D \interfields to form constraints for downstream applications. 

\noindent
\qheading{Neural Reconstruction (\cref{subsec:lexisflow,subsec:method:refine})}
We build \lexisFlow (\cref{subsec:lexisflow}), a dual-stream Flow-Matching transformer, that takes an image (and object shape estimation) and estimates a 3D human and object in interaction, along with a \lexisDict code describing this. 
This \lexisDict code is decoded into \interfields to guide sampling to refine 
(\cref{subsec:method:refine}) body and object pose, along with updated \interfields, in a form of neural analysis-by-synthesis. 
This makes \lexisFlow a \emph{one-stage end-to-end} model, which does not require post-hoc optimization. 
    
\subsection{\lexisDict: Learned Interaction Signatures}
\label{subsec:lexisnet}

\hspace{\parindent}
\interfields store a distance value for every point across a 3D surface (to the closest point on the interacting counterpart).
Thus, they are high-dimensional ($\nR^{6890}$ for \SMPLH), so they are challenging to infer. 
Moreover, estimating 3D \interfields from only a single 2D image is highly ill-posed. 
Our key observation is that \interfields encode action- and object-specific patterns, \ie, compact ``interaction signatures''. 
Thus, these can be used to infer low-dimensional interaction signatures, which can be decoded into full \interfields to form geometric losses. 

To this end, we employ a \VQVAE, called \textbf{\lexisNet}, to learn a novel compact \emph{lexicon of interaction signatures}, called \textbf{\lexisDict}; see \cref{fig:lexisnet} for an overview. 
\lexisNet 
builds on the pose tokenization of \tokenHMR~\cite{dwivedi_cvpr2024_tokenhmr} and extends it so that, conditioned on object shape in canonical pose (denoted as point cloud $\objPC \in \nR^{N \times 3}$), a decoder reconstructs both 3D body pose and 3D (body and object) \interfields. 
Thus, the learned tokens encode not only the body configuration, but, crucially, also the proximal human-object relationships and physical coupling that describes the interaction computationally. 

In detail, an encoder $\cE_\phi$ (with network weights $\phi$) maps body pose, $\pose \in \nR^{\numJoints \times 3}$ (axis-angle 
rotations for 21 joints up to the wrist), to a discrete token sequence (one token per joint), $\lexisTokens \in \nR^{\lexisDim}$, via a learned codebook, $\codebook = \{\centry_k\}_{k=1}^{K}$, $\centry_k \in \nR^{\lexisDim}$, called \textbf{\lexisDict} (for \lexisDictLONG):
\begin{equation}
    \lexisTokens = \arg\min\nolimits_{\centry_k \in \codebook} \|\zhat - \centry_k\|_2,     \qquad
    \zhat = E_\phi(\pose), 
    \label{eq:vqvae_encode}
\end{equation}
where 
$\zhat$         is a sequence of continuous latents, and 
$\lexisTokens$  of discrete latents, 
                the quantization operator ($\operatorname*{arg\,min}$) maps continuous latents ($\zhat$) to their nearest discrete codebook entry ($\centry_k$), and 
$D$             is the token length and 
$K$             the codebook size; 
                for the values of the latter two, see ``Implementation Details'' in \cref{subsec:implementation_details}.

The decoder \lexisDecoder (with network weights $\psi$) maps latents $\lexisTokens$ to full \interfields. 
Naively doing so for a specific mesh is straightforward, but ties the decoder, \lexisDecoder, to a specific topology; 
note that different body models have a different number of vertices, and objects vary widely. 
To tackle this, \lexisDecoder projects quantized tokens $\cZ$ into two TriPlane~\cite{chan2022eg3d} feature volumes: 

\input{fig/04_triplane}

\noindent
\vspace{-0.5 em}
\begin{equation}
    \triplaneHum,\, \triplaneObj = \lexisDecoder(\lexisTokens,\, \objfeat), 
    \label{eq:triplane_decode}
    \vspace{+0.5 em}
\end{equation}
where $\triplaneHum, \triplaneObj \in \nR^{\triplaneDim}$ are TriPlane feature volumes for the body and object respectively, and 
$\objfeat$ is an object shape embedding extracted by a pretrained frozen \mbox{PointNeXt}~\cite{pointnext} encoder applied on the object point cloud, $\objPC$. 
To query the decoded \interfield at any 3D surface point, $\querypt \in \nR^3$, \lexisNet projects $\querypt$ onto the three axis-aligned planes of the respective TriPlane (see \cref{fig:triplane}), bilinearly samples and sums the features, and infers an \interfield value through a lightweight MLP.

The \lexisNet training objective is:
\begin{equation}
    \cL_{\text{LexisNet}} =                                 \cL_{\text{recon}}  + 
                                \interFieldLossWeight^{h}\, \interFieldLoss^{h} + 
                                \interFieldLossWeight^{o}\, \interFieldLoss^{o} + 
                                \lambda_{\text{VQ}}\,       \cL_{\text{VQ}}     + 
                                \lambda_{\text{commit}}\,   \cL_{\text{commit}}
                                \text{,}
    \label{eq:lexisnet_loss}
\end{equation}
where 
$\cL_{\text{recon}}$        is a reconstruction loss defined as $\cL_{\text{recon}} = \big\| \cM(\posePred,\shape) - \cM(\poseGt,\shape) \big\|_2$, 
$\cM(\cdot,\cdot)$                denotes the \smplH \cite{MANO:SIGGRAPHASIA:2017} body mesh produced for given pose and shape parameters, 
$\posePred$ and $\poseGt$   are the estimated and GT body poses, respectively, $\shape$ is GT \smplh shape parameters, and 
$\interFieldLoss$           is a loss that 
                            penalizes the discrepancy between estimated and GT \interfields, 
                            $\interFieldPRED(\querypt)$ and $\interFieldGT(\querypt)$, respectively, as 
                            $\interFieldLoss = \frac{1}{|\surfpts|}\sum_{\querypt \in \surfpts} |\interFieldPRED(\querypt) - \interFieldGT(\querypt)|$, 
                            where $\querypt$ are 1024 points uniformly sampled on a surface $\surfpts$; 
                            we compute $\interFieldLoss$ separately for the body ($\interFieldLoss^\humSuper$) and object ($\interFieldLoss^{o}$).
The codebook losses $\cL_{\text{VQ}}$ and $\cL_{\text{commit}}$ follow standard \VQVAE training~\cite{dwivedi_cvpr2024_tokenhmr}. 
The steering weights $\lambda$ are set empirically; see ``Implementation Details'' in \cref{subsec:implementation_details}.

In \cref{subsec:lexisflow,subsec:method:refine} 
we refer to the codebook $\codebook$ as ``\textbf{\lexisDict}'', and to the \emph{continuous} $\zhat$ as ``\textbf{\lexisDictCode}'' as diffusion operates on codes $\zhat$. 
In practice, estimated $\interFieldPRED{=}\lexisDecoder(\lexisTokensPRED,\objfeat)$ is mapped via $\exp(-\omega \cdot \interFieldPRED)$ to emphasize close proximity.

\vfill
\pagebreak

\input{fig/05_lexisdiff}

\subsection{\lexisFlow}
\label{subsec:lexisflow}

\hspace{\parindent}
After pretraining a \lexisDict codebook of interaction signatures (\cref{subsec:lexisnet}), we exploit this to reconstruct 3D \HOI from an image. 
To this end, we build \lexisFlow (see \cref{fig:lexisdiff} for an overview), a model that captures the conditional distribution \mbox{$P(\humState, \objState \mid \img)$} via Flow Matching~\cite{flowmatching2023lipman}. 
The generative process is formulated as an Ordinary Differential Equation (\ODE) that transports samples from a Gaussian prior $\cN(\bzero, \bI)$ at $t{=}0$ to the target \HOI distribution at $t{=}1$.

Let $\hoiState = [\humState;\, \objState]$ denote the HOI state.
Then, the flow interpolates between Gaussian noise $\noise \sim \cN(\bzero, \bI)$ at $t{=}0$ and a ground-truth sample $\hoiState_1$ at $t{=}1$ via $\hoiState_t = (1 - t)\,\noise + t\,\hoiState_1$.
To approximate the vector field that generates this flow, we train a neural network, $\velfield(\hoiState_t, t, \img)$; then the target flow velocity is $\nicefrac{d\hoiState_t}{dt} = \hoiState_1 - \noise$.

The vector field, $\velfield$, is parameterized by a multi-stream transformer backbone inspired by DiT~\cite{peebles2023dit} and Mixture of Transformers~\cite{SAM3D::2025}.
The body stream (for state $\humState$), and object stream (for state $\objState$), operate with decoupled noise schedules, conditioned on independent timesteps $t_1$ and $t_2$, respectively, as in \mbox{TriDi}~\cite{petrov2025tridi}.
Cross-attention between the two streams conditions each modality on the other one, with sinusoidal timestep embeddings informing the network of each stream's noise level.
During training, $t_1$ and $t_2$ are sampled independently~\cite{bao2023unidiffuser}.
This lets the network model both the joint distribution $P(\humState, \objState \mid \img)$ and the conditional ones, $P(\humState \mid \objState, \img)$ and $P(\objState \mid \humState, \img)$, preventing modality collapse (where one stream might get ignored) 
by requiring each stream to be informative at all noise levels.

The \lexisFlow training objective is:
\begin{equation}
    \cL_{\text{\lexisFlow}} = \cL_{\text{FM}} +  
                                \interFieldLossWeight\, \interFieldLoss     + 
                                \lambda_{\text{v2v}}\,  \cL_{\text{v2v}}    + 
                                \lambda_{\text{2D}}\,   \cL_{\text{2D}}     + 
                                \lambda_{\text{obs}}\,  \cL_{\text{obs}}
                        \text{,}
    \label{eq:lexisflow_loss}
\end{equation}
where $\cL_{\text{FM}}$ is the conditional flow-matching loss:
\begin{equation}
    \cL_{\text{FM}} = \mathbb{E}_{t_1, t_2 \sim \mathcal{U}(0,1),\; \noise \sim \cN(\bzero,\bI)} 
                            \big\|
                                        \velfield(\hoiState_{(t_1, t_2)},\, t_1,\, t_2,\, \img) - \vtarget 
                            \big\|_2^2
                        \text{,}
    \label{eq:flow_matching_loss}
\end{equation}
where
$\hoiState_{t_1, t_2}$ is the noised state with body components interpolated at $t_1$ and object components at $t_2$, and 
$\vtarget$ is the conditional vector field target; note that the expectation is approximated by mini-batch averaging.
The rest of the loss terms provide explicit geometric and visual supervision; 
$\interFieldLoss$   penalizes the $L_1$ discrepancy between                     the estimated ($\lexisDecoder(\lexisTokensPRED, \objfeat)$) and GT \interfields, $\cL_{\text{v2v}}$ is a $L_2$  vertex-to-vertex loss between estimated ($\cM(\hat{\humState})$) and GT ($\cM(\humState)$) body meshes, 
$\cL_{\text{2D}}$ is a 2D keypoint \blue{re-projection} loss encouraging alignment with image cues, and $\cL_{\text{obs}}$ is a loss penalizing deviation between ``observed'' and GT \interfields, encouraging consistency between the estimated poses and \interfields. 
To this end, it first poses the object relative to the body using the estimated transformations $\{ \hat{\objR}, \hat{\objt}\}$, 
computes the ``observed'' \interfields between the posed body and object meshes, and penalizes their deviation from the GT \interfield. 
The steering weights $\lambda$ are set empirically; see ``Implementation Details'' in \cref{subsec:implementation_details}. 

\subsection{\magenta{\interfield-Guided Refinement} within \lexisFlow} 
\label{subsec:method:refine}

\hspace{\parindent}
Our \lexisFlow (\cref{subsec:lexisflow}) is a flow-based generative model, so it imposes a prior that pulls estimations towards the training-data distribution. 
However, for rare/unseen images and interactions it can naturally make errors. 
Typically, optimization performs corrections via contact constraints~\cite{xie2022chore}, but is slow, prone to local minima, and takes place post-hoc, \ie, in a \emph{separate} stage \emph{after} inference. 

We tackle this limitation with a \emph{guided refinement}; 
\lexisFlow \emph{itself} performs the refinement in inference time (\emph{without} additional stages) by adding gradient-based guidance steps in the \ODE sampling loop. 
Specifically, at intermediate timesteps $t$, the \ODE solver pauses, and the current state $\hoiState_t$ is updated via a guidance gradient before the next iteration step. 

To this end, we add a guidance loss with the terms discussed in the following: 
\begin{equation}
    \mathcal{L}_{\text{guide}} = \interFieldGuideLoss + \maskWeight \mathcal{L}_{\text{mask}}
    \text{.}
\end{equation}

\zheading{\interfield Guidance ($\interFieldLoss$)} 
The \lexisNet decoder, $\lexisDecoder$, maps the running estimations of latent tokens, $\lexisTokensPRED$, to a predicted \interfield, $\interFieldPRED = \lexisDecoder(\lexisTokensPRED, \objfeat)$, which 
encodes the proximal relationships between interacting surfaces. 
Moreover, an ``observed'' \interfield is computed based on the running estimations for the body and object meshes. 
Then, the novel loss term $\interFieldGuideLoss$ penalizes the discrepancy between the ``observed'' and estimated \interfields: 
\begin{equation}
    \interFieldGuideLoss = \| \interFieldPRED - \interFieldGT_t\|^2
    \text{,}
\end{equation}
where $\interFieldGT_t$ denotes the computed \interfield by posing body (decoded from running $\lexisTokensPRED$) and object with running estimates $\{\humR_t, \humt_t\}, \{\objR_t, \objt_t\}$, respectively.
This acts as a field that pulls the object and body into a 3D configuration informed by the \lexisDict interaction tokens and their corresponding \interfields, in a form of \lexisDict-/\interfield-based analysis-by-synthesis. 

\zheading{Mask-pixel Guidance ($\mathcal{L}_{\text{mask}}$)} 
To align 3D estimations with image cues, a common way is to use mask-based losses. 
Doing this for the human is challenging, as people have hair, accessories, and loose clothing that ``contaminate'' masks and cannot be explained by parametric body models such as \smplH. 
However, doing this for the object is reliable, because the recent SAM \cite{SAM:ICCV:2023} model estimates robust 2D masks, and SAM3D \cite{SAM3D::2025} provides good 3D shape estimates (though pose estimates are noisy). 
Thus, we add a mask-based loss for the object. 

Object vertices $\objverts$, transformed by \blue{running state estimate} $(\objR, \objt)$, are projected \blue{onto the image plane} using a differentiable renderer~\cite{ravi2020pytorch3d}.
Vertices falling outside the observed 2D object mask~\cite{SAM:ICCV:2023}, $\objmask$, are penalized by:
\begin{equation}
    \cL_{\text{mask}} = \sum\nolimits_{\bv \in \objverts}   \| 1 - \objmask
                                                                    \big( 
                                                                        \proj ( \objR_t \bv + \objt_t )
                                                                    \big) 
                                                            \|^2  
                                                            \text{,}
    \label{eq:mask_guidance}
\end{equation}
providing a gradient signal that aligns the 3D shape with the 2D image cues.

\zheading{Guided Flow Trajectory}
The state $\hoiState_t$ is updated by a gradient step:
\begin{equation}
    \hoiState_t \leftarrow \hoiState_t - \guidestep \cdot \nabla_{\hoiState_t} (\interFieldGuideLoss + \maskWeight \cL_{\text{mask}}\, ).
    \label{eq:guidance_update}
\end{equation}
Correcting $\hoiState_t$ during sampling leads to a \emph{\blue{guided flow} trajectory}, dynamically balancing interaction awareness (\lexisDict-based \interfields) and image cues (mask).
This update performs guided refinement, \emph{without} additional, separate stages. 

Importantly, this formulation also lets \lexisFlow exploit off-the-shelf \sota 3D body-only and object-only estimators that are \emph{not} interaction-aware, so that it has a better initialization for adding the missing interaction awareness. 
That is, given rough initial 3D body and object estimations from a baseline model, \lexisFlow encodes these into a unified latent state $\hoiState_{\text{init}} = \{\humState_{\text{init}}, \objState_{\text{init}}\}$.
Then, rather than generating \blue{a 3D \HOI estimation} from pure noise, a controlled amount of noise is injected to transport this estimate to an \emph{intermediate} flow timestep $t_{\text{start}} \in (0, 1)$:
$\hoiState_{t_{\text{start}}} = (1 - t_{\text{start}})\,\noise + t_{\text{start}}\,\hoiState_{\text{init}}$, as in \mbox{SDEdit} \cite{meng2022sdedit}.
A partial ODE integration then proceeds from $t_{\text{start}}$ to $t{=}1$ with the guided sampling.
In the ``Experiments''section (\cref{sec:experiments}), this variant is denoted as \lexisFlowRefinement. 

\subsection{Implementation Details}
\label{subsec:implementation_details}

\hspace{\parindent}
\zheading{\lexisDict}
We initialize \lexisNet VQ-VAE with \tokenHMR~\cite{dwivedi_cvpr2024_tokenhmr} tokenizer and further train it to \interfields (evaluated at 1024 uniform surface points).
We use $\omega{=}5$ for exponential mapping of \interfields.
The \lexisDict codebook contains $K{=}2048$ entries ($D{=}128$). 
A frozen PointNeXt~\cite{pointnext} (pretrained on ModelNet40~\cite{shapenet}) encodes object geometry into a 1024-D global descriptor. 
This conditions the $\triplaneHum$ and $\triplaneObj$ TriPlanes ($\triplaneRes{=}64$ resolution, $\triplaneChan{=}64$ channels), which feed a 2-layer MLP ($[128, 64]$ hidden units). 
We optimize \cref{eq:lexisnet_loss} ($\interFieldLossWeight^{h}{=}10$, $\interFieldLossWeight^{o}{=}10$, $\lambda_{\text{VQ}}{=}\lambda_{\text{commit}}{=}0.5$) for $60{,}000$ iterations using AdamW~\cite{adamw} ($\text{lr}{=}10^{-4}$, batch size 1024). 
Training takes ${\sim}5$ hours on a single RTX-6000Ada. %

\zheading{\lexisFlow} 
\lexisFlow uses a 18-layer multi-stream transformer backbone \cite{SAM3D::2025} ($512$ hidden units). 
As image encoder we use frozen DINOv2 (\mbox{ViT}/\mbox{L-14}) \cite{oquab2024dinov2}.
Optimizing the training loss of \cref{eq:lexisflow_loss} 
($\interFieldLossWeight{=}10$, 
$\lambda_{\text{v2v}}{=}5$, 
$\lambda_{\text{2D}}{=}1$, 
$\lambda_{\text{obs}}{=}0.5$) 
runs for $100$k steps via AdamW~\cite{adamw} (batch size $1024$). 
We use a warmup cosine plateau scheduler (learning rate $10^{-6}$ to $5{\times}10^{-4}$) and a $0.3$ dropout rate for classifier-free guidance, with guidance-scale $1.5$. 
Training takes ${\sim}48$ hours on 4 RTX-6000Ada GPUs. 
We use Euler ODE solver ($25$ steps) for sampling and perform guidance/refinement (\cref{eq:guidance_update}) starting at $t_{start}{=}15$, $\lambda_{\text{mask}}{=}0.5$.

%% file: fig/03_lexisnet.tex
 \begin{figure*}[t]
    \centering
    \vspace{-1.0 em}
    \includegraphics[trim=000mm 000mm 000mm 000mm, clip=true, width=0.99 \textwidth]{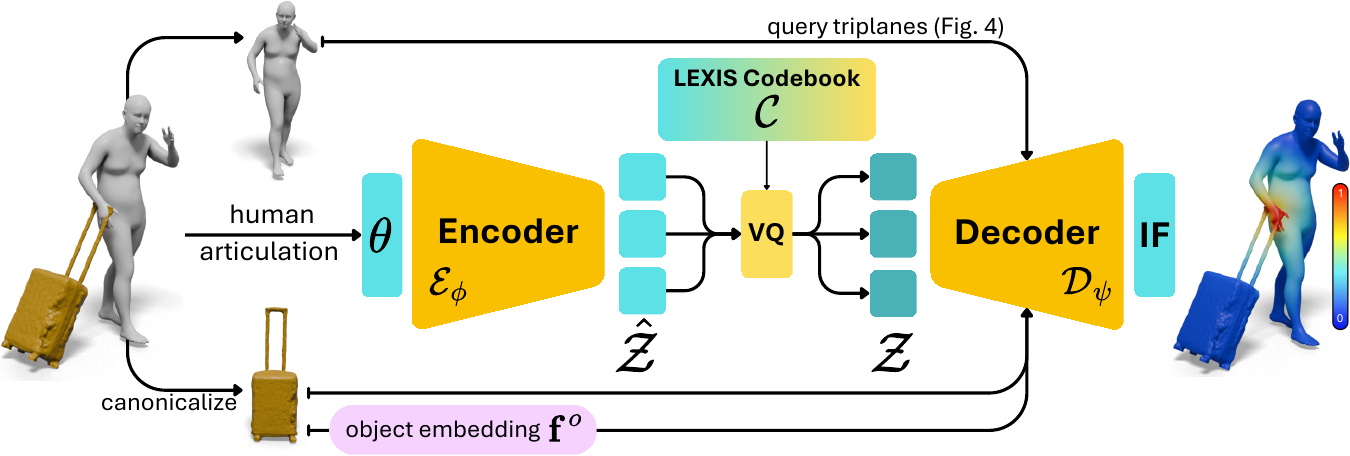}
    \vspace{-1.0 em}
    \caption{
            \qheadingFIGTAB{\lexisDict (\cref{subsec:lexisnet})} 
            To estimate 3D \HOI from images we need a prior model of interactions. 
            We train \lexisNet, a \VQVAE that learns a compact dictionary
            of proximal interaction signatures, termed \lexisDict. 
            The encoder $\cE_\phi$ maps 3D body pose to continuous latents $\zhat$, quantized via a learned codebook $\codebook$ into discrete tokens $\cZ$, and decoded via $\cD_\psi$ into 3D body pose and body/object \interfields (shown color-coded). 
    }
    \label{fig:lexisnet}
    \vspace{-0.5 em}
\end{figure*}

%% file: fig/04_triplane.tex
\begin{wrapfigure}{r}{0.46 \textwidth}
    \centering
    \vspace{-2.20 em}
    \includegraphics[width=0.45 \textwidth]{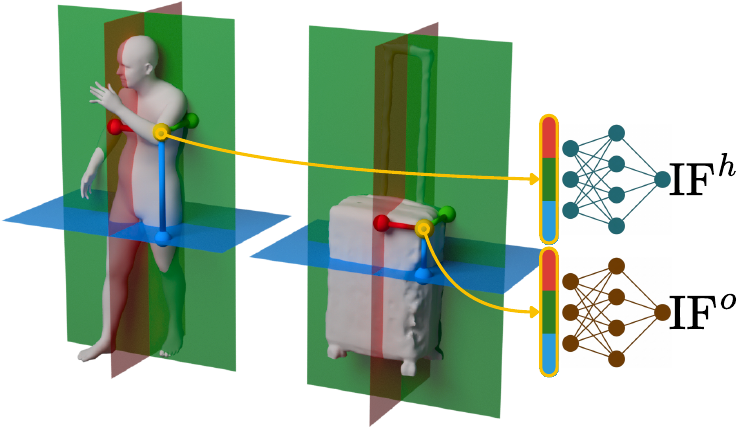}
    \vspace{-1.00 em}
    \caption{
                    \qheadingFIGTAB{TriPlane-based \interfields} 
                    A 3D surface point $\querypt$ (see yellow point on human/object surface) is orthogonally projected onto the three feature planes (see red, blue, and green points) to sample features, 
                    which are aggregated and passed to a small MLP to infer the \interfield value for point $\querypt$. 
    }
    \vspace{-1.0 em}
    \label{fig:triplane}
\end{wrapfigure}

%% file: fig/05_lexisdiff.tex
\begin{figure*}
    \centering
    \vspace{-0.5 em}
    \includegraphics[trim=000mm 000mm 000mm 000mm, clip=true, width=0.99 \textwidth]{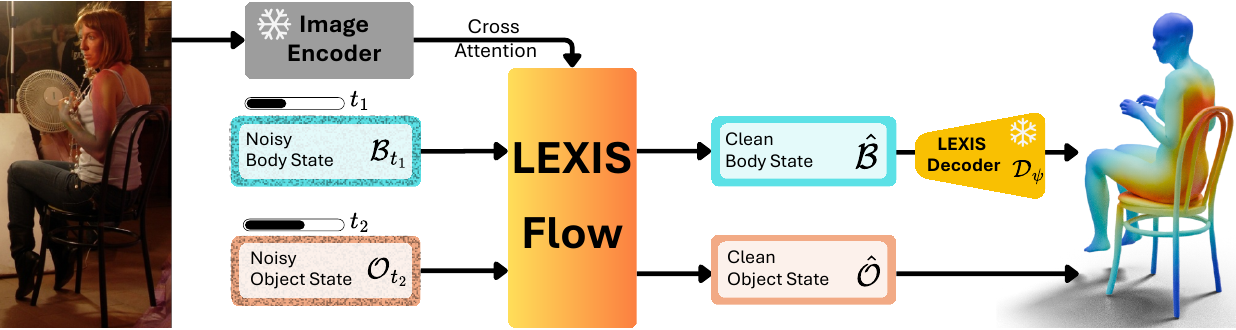}
    \vspace{-0.75 em}
    \caption{
            \qheadingFIGTAB{\lexisFlow (\cref{subsec:lexisflow})} 
            We develop a dual-stream Flow-Matching model that takes as input a single image, and estimates 3D body and object meshes in interaction along with \lexisDict-based \interfield 
            proximal relationships (\cref{subsec:lexisnet}; shown with heatmap color-coding on the 3D meshes).
            \blue{Guiding sampling} via \lexisDict-based \interfields refines estimates to improves the physical plausibility of 3D interaction.
    }
    \label{fig:lexisdiff}
    \vspace{-1.5 em}
\end{figure*}

%% file: sec/04_Experiments.tex
\section{Experiments}
\label{sec:experiments}
\subsection{Experimental Setup}
\label{subsec:exp_setup}

\hspace{\parindent}
\zheading{Training Data}
We train \lexisNet on geometry-only \HOI datasets, combining the synthetic \procigen~\cite{xie2024hdm_procigen} with \InterAct-refined~\cite{InterAct:CVPR:2025} \mocap datasets: \neuraldome~\cite{zhang2023neuraldome}, \omomo~\cite{li2023omomomo}, and \imhd~\cite{zhao2024imhoi}.
We train \lexisFlow on \procigen 3D-paired images for image-conditioned generation, and include 3D-only \InterAct-refined data for classifier-free guidance.

\zheading{Benchmarks}
We use the \inthewild \opentdhoi~\cite{open3dhoi} (\cref{subsec:itw_eval}) and the in-lab \behave~\cite{bhatnagar22behave} dataset (\cref{subsec:controlled_eval}). 
For the former we have 2 tasks; see below, where our generative model (sampling from noise) is denoted as \lexisFlow and a refinement variant (initialized by \sota models via \SDEdit~\cite{meng2022sdedit}) as $\lexisFlow^*$.

\zheading{Generative Reconstruction (\opentdhoi)} \lexisFlow estimates 3D \emph{directly from noise}, conditioned on image features, without any external initialization for body or object pose.
For \blue{fair} comparison, we train \lexisFlow only on \procigen, as the \hdm~\cite{xie2024hdm_procigen} baseline. 
For object shape, we provide the \mbox{SAM3D} shape estimate as input to \lexisFlow.
For HDM, we use its object-specific version for classes it was trained on, and its general version for all others.

\zheading{Guided Refinement (\opentdhoi)}
$\lexisFlow^*$ starts from off-the-shelf estimates rather than noise; 
it uses \camerahmr~\cite{camerahmr} for the body and \samtd~\cite{SAM3D::2025} for the object, aligned to a \mbox{MoGe}~\cite{wang2025moge} depth estimate; see alignment details in \cref{subsec:supmat:alignment}. 
This initialization is encoded into the latent state and transported to an intermediate timestep \blue{$t_{\text{start}}{=}15$ (out of total 25)} via \mbox{SDEdit}~\cite{meng2022sdedit}, after which InterField-guided \ODE integration refines estimates (\cref{subsec:method:refine}).
\lexisFlowRefinement trains on a combination of the \procigen and \InterAct-refined datasets.
We compare against the optimization-based \interactvlm~\cite{dwivedi_interactvlm_2025} and \hoigaussian~\cite{open3dhoi} methods, initialized with \camerahmr + \samtd + \mbox{MoGe} as our method for fair comparison. 
We refer to this version of \interactvlm as ``\interactvlmSAMthreeD.''
Originally, \interactvlm uses ICP-based initialization of object pose, which struggles in difficult scenarios; 
for completeness, we also report this original setting, denoted as ``\interactvlm.''

\zheading{In-lab Evaluation (\behave)}
For fair comparison to baselines that train on \behave, we train dataset-specific variants of \lexisNet and \lexisFlow on this dataset, using the official train/test split~\cite{xie2022chore}.
We evaluate under the generative setting only; \blue{for this}, \lexisFlow generates the \HOI state from noise without any external expert initialization.
We compare against \chore~\cite{xie2022chore}, \contho~\cite{nam2024contho}, \hoitg~\cite{wanghoi-tg}, and \hdm~\cite{xie2024hdm_procigen}, all trained on \behave.

\zheading{Evaluation Metrics}
We apply Procrustes alignment to the predicted body meshes and apply the same transformation to the object as in ~\cite{nam2024contho,wanghoi-tg}, before computing all metrics. 
To tackle the different topology of \smplx~\cite{SMPL-X:2019} (\opentdhoi) and \smplh~\cite{MANO:SIGGRAPHASIA:2017} (ours), we align their common vertices and evaluate on uniformly-sampled surface points. 
We report Chamfer Distances ($\text{CD}_{\text{hum}}$, $\text{CD}_{\text{obj}}$; cm, lower is better) for geometry, Collision (\% of penetrating vertices~\cite{open3dhoi}; lower is better) for physical plausibility, and Contact F1 (@5cm~\cite{nam2024contho}; higher is better) for interaction fidelity.
For metric formulas, see \cref{subsec:supmat:metrics}.

\input{tab/01_open3dhoi}

\input{tab/02_representation}

\subsection{Interaction Representation}
\label{subsec:representation_eval}

\hspace{\parindent}
To evaluate the effect of the interaction representation, we compare the dense, continuous \interfields with sparse, binary contacts; for fairness, contacts are extracted by thresholding \interfields to \blue{compute a level set}. 
We evaluate on two tasks: 
(1) The render-and-compare fitting of \interactvlm~\cite{dwivedi_interactvlm_2025}. 
(2) The guided generative sampling of \lexisFlow. 

The evaluation results are shown in \cref{tab:representation_eval}. 
For both tasks, the dense, continuous \interfields outperform the sparse, binary contacts. 
Specifically, for the ``fitting'' task, $\text{CD}_{\text{obj}}$ improves by 2.26\% and Contact by 12.24\%; for ``generative'' task the improvement is 16.49\% and 90.79\% respectively. 
This \blue{shows} that the dense, continuous \interfield representation encodes richer spatial signal than sparse, binary contacts, helping 3D reconstruction and human-object alignment. 

\subsection{\Inthewild 3D \HOI Reconstruction}
\label{subsec:itw_eval}

We evaluate 3D \HOI estimation on the \inthewild images of \opentdhoi~\cite{open3dhoi}; 
we evaluate two \lexisFlow versions, for ``generative reconstruction,'' and for ``guided refinement.'' 
The results are shown in \cref{tab:open3dhoi_benchmark} and \cref{fig:results_itw}. 

\zheading{\blue{Generative Reconstruction (\cref{tab:open3dhoi_benchmark})}} 
\lexisFlow~\magenta{(row~B)} outperforms \hdm~\cite{xie2024hdm_procigen}~\magenta{(row~A)} on $\text{CD}_{\text{hum}}$ (8.8 vs\ 13.5\,cm), $\text{CD}_{\text{obj}}$ (35.0 vs\ 49.3\,cm) and Contact F1 (0.21 vs\ 0.14);
note in \cref{fig:results_itw} that \hdm produces geometric artifacts and intersecting meshes on real-world images. 
Notably, \lexisFlow~\magenta{(row~B)}, with its direct generative estimation, already achieves lower $\text{CD}_{\text{obj}}$ and higher Contact F1 than a baseline that combines off-the-shelf \camerahmr + \samtd{}~(row~D) estimations (aligned using MoGe~\cite{wang2025moge} depth). 
This suggests that jointly estimating humans and objects by exploiting interaction signatures (\interfields)~(row~B) 
produces better spatial configurations than independently combining per-entity experts~(row~D). 
When we initialize our method with these off-the-shelf estimates~(row~D), denoted as $\lexisFlowRefinement$(row~G), it improves further; see below.

\input{fig/06_results_itw_vs_sota}

\zheading{Guided Refinement (\cref{tab:open3dhoi_benchmark})}
With identical initialization~\magenta{(row~D--G)} $\lexisFlow^*$~\magenta{(row~G)} achieves the best results across all four metrics, outperforming both \interactvlm~\magenta{(row~C)} and \hoigaussian~\magenta{(row~E)}.
$\lexisFlowRefinement$ also outperforms \interactvlmSAMthreeD~\magenta{(row~F)} (initialized with \camerahmr + \samtd + MoGe; see \cref{subsec:exp_setup}).
This shows that the gains are not solely due to better initialization; 
even when \interactvlm benefits from the same starting point, the InterField-guided refinement of $\lexisFlowRefinement$ yields stronger spatial alignment.
The key difference is that the optimization baselines refine in output space with sparse contact losses, while $\lexisFlowRefinement$ corrects intermediate latent states during generation using dense \interfield signals from the learned \lexisDict. 

\zheading{Qualitative comparison}
\Cref{fig:results_itw} compares our $\lexisFlowRefinement$ method against \sota methods in the wild; for more results see \cref{subsec:supmat:additional_results}.
Optimization-based methods often produce floating objects or penetration, because sparse contacts lack the spatial awareness to correct these errors. 
$\lexisFlowRefinement$ recovers physically-plausible interactions via \interfield-guided refinement. 
Additional results in \cref{fig:results_additional} show our \itw generalization; for more results see \cref{subsec:supmat:additional_results}.

\zheading{Perceptual study}
We conduct a perceptual study on 60 Open3DHOI images with 62 participants (protocol in \cref{subsec:supmat:perceptual}). 
This shows that $\lexisFlowRefinement$ estimations are perceived as more realistic 75.8\% of times over \hoigaussian.

\input{tab/04_sota_behave}

\subsection{In-lab 3D HOI Reconstruction}
\label{subsec:controlled_eval}

We evaluate in-lab HOI reconstruction on the \BEHAVE benchmark (\cref{tab:compare_sota_behave_intercap}).
\lexisFlow resets the \sota performance on the \BEHAVE benchmark by a significant margin of 15.0\% and 5.0\% on CD$_{\text{hum}}$ and CD$_{\text{obj}}$ respectively relative to the closest competitor, \hoitg~\cite{wanghoi-tg}.

\subsection{Ablation Study}
\label{subsec:ablation}

\input{tab/03_guidance_ablation}

\input{fig/07_results_additional}

\Cref{tab:ablation} evaluates our design choices on the Open3DHOI~\cite{open3dhoi} dataset.

\zheading{Architecture (unguided)}
When replacing the discrete \lexisDict codebook with a continuous VAE latent
(\textit{GaussFlow}) degrades $\text{CD}_\text{hum}$ by
39\% and $\text{CD}_\text{obj}$ by 19\%. 
This shows that discrete tokenization preserves pose expressivity and interaction structure.

After removing \lexisDict, 
we keep the human fixed (initialized with the \sota CameraHMR method) and denoise object pose (\emph{\flowPose}). 
This worsens performance, showing that jointly denoising the body and object with \lexisDict is essential; 
note that $\text{CD}_\text{hum}$ is not reported as body parameters are fixed.

\zheading{Guidance signal (\cref{subsec:method:refine})}
Adding mask guidance ($\mathcal{L}_{\text{mask}}$) reduces $\text{CD}_\text{obj}$ to $43.51$.
InterField guidance ($\interFieldGuideLoss$) achieves $\text{CD}_\text{obj}$ of $41.01$. 
Combining both yields the best $\text{CD}_\text{obj}$ (35.01), a 27\%
reduction over unguided sampling. 
This shows that 2D masks and 3D \interfield proximity are complementary signals.

%% file: tab/01_open3dhoi.tex
\begin{table}[t]
    \centering
    \vspace{-0.5 em}
    \resizebox{ \columnwidth}{!}{
        \begin{tabular}{lcccc}
            \toprule

             Method & \multicolumn{1}{l}{CD$_{\text{hum}} \downarrow$} & \multicolumn{1}{l}{CD$_{\text{obj}} \downarrow$} & \multicolumn{1}{l}{Collision $\downarrow$} & \multicolumn{1}{l}{Contact $\uparrow$} \\
            \midrule

            \magenta{A.}~\hdm~(w.~scale align.)~\cite{xie2024hdm_procigen}
            & \metricbetter{13.50}{34.4} & \metricbetter{49.38}{29.1} & \metricbetter{0.089}{32.6} & \metricbetter{0.141}{49.6} 
            
            \\
            \magenta{B.}~\lexisFlow (Ours) & \textbf{8.85}~\makebox[3.8em][l]{} & 
            \textbf{35.01}~\makebox[3.9em][l]{} & 0.060~\makebox[4.em][l]{} & 0.211~\makebox[4.em][l]{} 
            \\

            \midrule
            
            \magenta{C.}~\interactvlm~\cite{dwivedi_interactvlm_2025} & \metricbetter{7.20}{2.1} & \metricbetter{38.20}{39.9} & \metricbetter{0.054}{24.1} & \metricbetter{0.372}{21.2} 
            \\
            \magenta{D.}~\camerahmr\cite{camerahmr} + \samtd~\cite{SAM3D::2025} & \metricbetter{7.20}{2.1} & \metricbetter{37.30}{38.4} & \metricbetter{0.051}{19.6} & \metricbetter{0.182}{147} 
            \\
            \magenta{E.}~\hoigaussian~\cite{open3dhoi} & \metricbetter{7.28}{3.2} & \metricbetter{32.02}{28.3} & \metricbetter{0.061}{32.8} & \metricbetter{0.151}{198} 
            \\
            \magenta{F.}~\interactvlmSAMthreeD~\cite{dwivedi_interactvlm_2025} & \metricbetter{7.20}{2.1} & \metricbetter{30.11}{23.7} & \metricbetter{0.047}{12.8} & \metricbetter{0.394}{14.5} 
            \\
            \magenta{G.}~\lexisFlowRefinement (Ours) & \textbf{7.05}~\makebox[4.em][l]{} & \textbf{22.96}~\makebox[4.em][l]{} & \textbf{0.041}~\makebox[3.9em][l]{} & \textbf{0.451}~\makebox[3.9em][l]{} 
            \\
            \bottomrule
        \end{tabular}
    }
    \caption{
        \qheadingFIGTAB{3D \HOI in the wild (\cref{subsec:itw_eval})} 
        We evaluate our \lexisFlow against \sota refinement-based methods 
        on the \opentdhoi~\cite{open3dhoi} benchmark.
        \magenta{Rows A--B} estimate from scratch; \magenta{row C} initializes with CameraHMR, and \magenta{rows E--G} initialize with \magenta{D}.    }
    \label{tab:open3dhoi_benchmark}
    \vspace{-2.0 em}
\end{table}

%% file: tab/02_representation.tex
\begin{wraptable}{r}{0.54 \textwidth} %
    \centering
    \vspace{-2.4 em}
    \resizebox{0.50 \columnwidth}{!}{
        \begin{tabular}{llcc}
            \toprule
            Task        & Representation      & CD$_{\text{obj}}$ $\downarrow$ & Contact $\uparrow$ \\
            \midrule
            Fitting     & Binary Contact      & 35.4            & 17.8 \\
            Fitting     & \interfield (ours)  & \textbf{34.6}   & \textbf{19.8} \\
            \midrule
            Generative  & Binary Contact      & 49.11           & 0.152 \\
            Generative  & \interfield (ours)  & \textbf{41.01}  & \textbf{0.290} \\
            \bottomrule
        \end{tabular}
    }
    \vspace{-0.5 em}
    \caption{
        \qheadingFIGTAB{Representation effect (\cref{subsec:representation_eval})}
        Contacts and \interfields are compared on \opentdhoi~\cite{open3dhoi} on two tasks: render-and-compare fitting and generative reconstruction.
    }
    \label{tab:representation_eval}
    \vspace{-1.5 em}
\end{wraptable}

%% file: fig/06_results_itw_vs_sota.tex
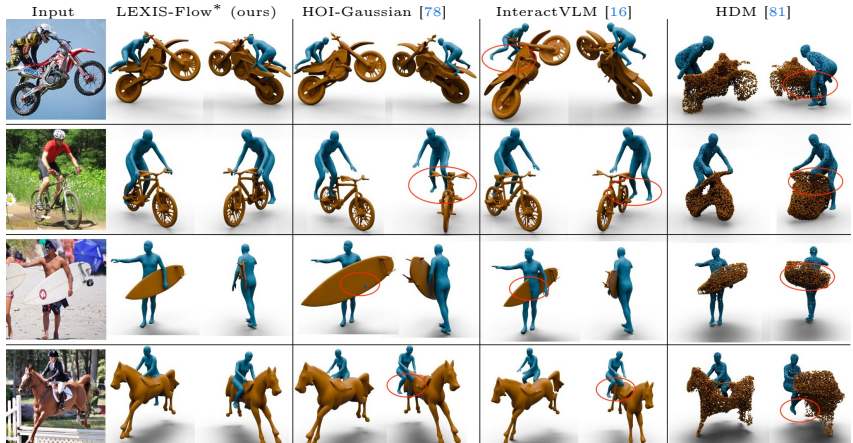
\begin{figure}
    \centering
    \begin{overpic}[width=0.92 \columnwidth, unit=1mm]{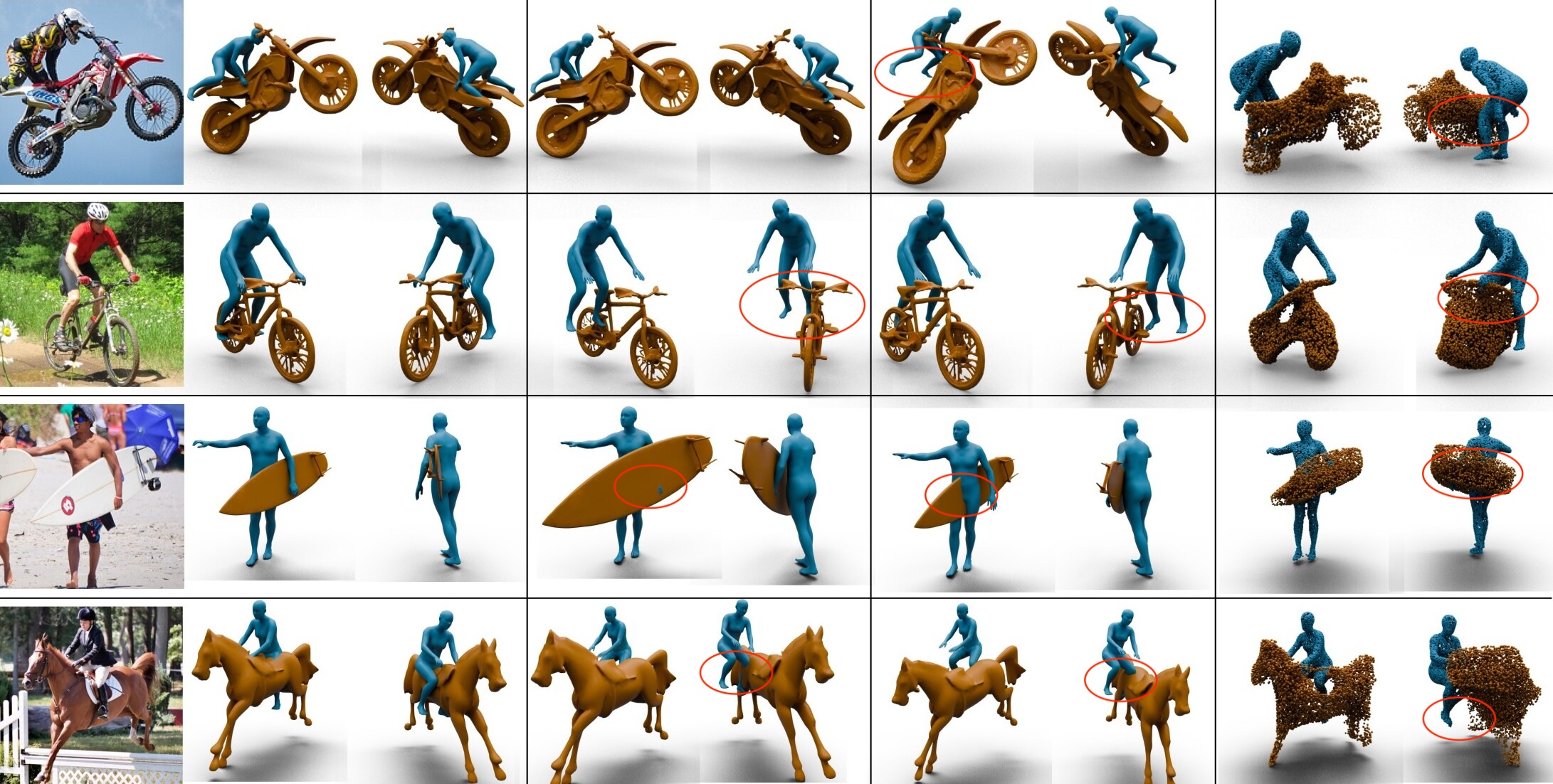}
        \put(3, 51){\tiny Input}
        \put(13, 51){\tiny $\lexisFlow^*$ (ours)}
        \put(35, 51){\tiny \hoigaussian~\cite{open3dhoi}}
        \put(58, 51){\tiny \interactvlm~\cite{dwivedi_interactvlm_2025}}
        \put(84, 51){\tiny \hdm~\cite{xie2024hdm_procigen}}
    \end{overpic}
    \vspace{-1.0 em}
    \caption{
            \qheadingFIGTAB{\lexisFlow vs \sota} 
            Existing methods often fail to capture tight physical coupling, yielding floating objects (\hoigaussian, \interactvlm) or penetrations (\hdm). 
            \lexisFlow tackles this via dense \interfields for proximity-aware estimation.
    }
    \label{fig:results_itw}
    \vspace{-1.0 em}
\end{figure}

%% file: tab/04_sota_behave.tex
\begin{wraptable}{r}{0.46 \textwidth}
    \centering
    \vspace{-5.2 em}
    \resizebox{0.43 \columnwidth}{!}{%
        \begin{tabular}{lrr}
            \toprule  
            \multirow{2}{*}{} & \multicolumn{2}{c}{\BEHAVE}  \\
            \cmidrule(lr){2-3} 
            & \multicolumn{1}{c}{CD$_{\text{hum}} \downarrow$} & \multicolumn{1}{c}{CD$_{\text{obj}} \downarrow$} \\
            \midrule
            \phosa~\cite{zhang2020phosa} & \metricbetter{12.17}{68.0} & \metricbetter{26.62}{71.5} \\
            \hdm~\cite{xie2024hdm_procigen} & \metricbetter{11.61}{66.4} & \metricbetter{11.35}{33.0}  \\
            \chore~\cite{xie2022chore} & \metricbetter{5.58}{30.1} & \metricbetter{10.66}{28.7} \\
            \contho~\cite{nam2024contho} & \metricbetter{4.99}{21.8} & \metricbetter{8.42}{9.7} \\
            \hoitg~\cite{wanghoi-tg} & \metricbetter{4.59}{15.0} & \metricbetter{8.00}{5.0} \\
            \midrule
            \lexisFlow & \textbf{3.90}~\makebox[3.8em][l]{} & \textbf{7.60}~\makebox[3.8em][l]{}  \\
            \bottomrule
        \end{tabular}
        }
    \vspace{-0.5 em}
    \caption{
                    \qheading{In-lab 3D \HOI}
                    Evaluation on the in-lab \BEHAVE~\cite{bhatnagar22behave} dataset.
    }
    \label{tab:compare_sota_behave_intercap}
    \vspace{-2.0 em}
\end{wraptable}

%% file: tab/03_guidance_ablation.tex
\begin{wraptable}{r}{0.50\textwidth}
    \vspace{-4.0 em}
    \centering
    \scriptsize
    \resizebox{\linewidth}{!}{ 
        \begin{tabular}{l|>{\centering\arraybackslash}p{2.7cm}|cc}
            \toprule
            & Variants & CD$_{hum} \downarrow$ & CD$_{obj} \downarrow$ \\
            \midrule
            \multirow{3}{1.5cm}{Architect. \\ (Unguided)} 
                & \lexisFlowGaussian  Baseline & 13.45 & 57.25 \\
                & \flowPose           Baseline & \NA & 65.05 \\
                & \lexisFlow          Baseline & \textbf{9.68} & \textbf{48.01} \\
            \midrule
            \multirow{4}{*}{Guidance}
                & Unguided                             & 9.68 & 48.01 \\
                & $\cL_{\text{mask}}$                  & 9.46 & 43.51 \\
                & $\interFieldGuideLoss$               & 9.05 & 41.01 \\
                & $\cL_{\text{mask}}$ + $\interFieldGuideLoss$ & \textbf{8.85} & \textbf{35.01} \\
            \bottomrule
        \end{tabular}
    }
    \vspace{-0.5em}
    \caption{
                \qheading{Ablations}
                We evaluate design choices for our architecture (unguided) and for guided refinement on \opentdhoi~\cite{open3dhoi}.
    }
    \label{tab:ablation}
    \vspace{-2.0 em}
\end{wraptable}

%% file: fig/07_results_additional.tex
\begin{figure}[t]
    \centering
    \vspace{-0.5 em}
    \includegraphics[width=0.97 \columnwidth]{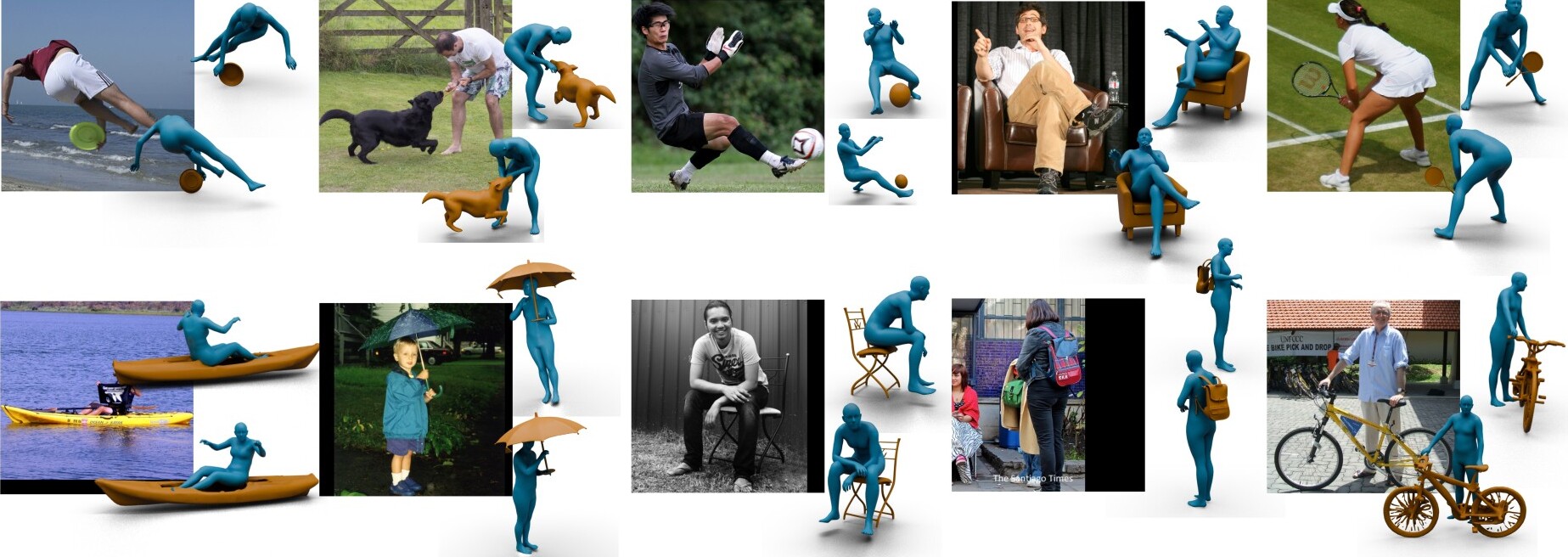}
    \vspace{-1.0 em}
    \caption{
        \qheadingFIGTAB{Qualitative results} 
        \lexisFlowRefinement recovers physically-plausible         interactions from diverse \inthewild images. 
        By leveraging dense \interfield signals, our model produces accurate spatial configurations and realistic articulations even under occlusion.
    }
    \label{fig:results_additional}
    \vspace{-1.0 em}
\end{figure}

%% file: sec/05_Conclusion.tex
\section{Conclusion}
\label{sec:conclusion}

We go beyond sparse, binary contact by leveraging dense, continuous \interfields for reconstructing 3D Human-Object Interaction from single images. 
To make InterField inference tractable from a single image, we learn \lexisDict, a latent manifold of interaction signatures, encoding interaction- and object-specific proximity patterns. 
Then, we develop \lexisFlow, a dual-stream Flow-Matching model that jointly estimates 3D human and object meshes alongside their \interfields, and exploits these for guided refinement, eliminating post-hoc optimization. 
Experiments show that continuous \interfields outperform binary contacts in both fitting and generative settings. 
On the \opentdhoi benchmark, \lexisFlowRefinement achieves the lowest $\text{CD}_\text{obj}$ (22.96) and highest Contact F1 (0.451), outperforming all baselines. 
This moves us closer to holistic 3D scene understanding. 

%% file: sec/06_Acknowledgments.tex
\section*{Acknowledgments}

We thank Bo\v{z}idar Anti\'{c} and Ilya Petrov for valuable insights and discussions. 
SKD is supported by the International Max Planck Research School for Intelligent Systems \mbox{(IMPRS-IS)}. 
We acknowledge HPC support by the \mbox{EuroHPC} Joint Undertaking that awarded access to the EuroHPC supercomputers \mbox{LEONARDO} (project ID \mbox{EHPC-AI-2024A06-077}), hosted by CINECA in Italy, and \mbox{JUPITER} (project ID \mbox{e-reg-2025r02-393}), hosted by JSC in Germany, and by the Dutch national e-infrastructure through the SURF Cooperative grant \mbox{no.~EINF-12852}. 
We also acknowledge support through a research gift from Google, and the NVIDIA Academic Grant Program.
This work is supported by the European Research Council (ERC) through the Starting Grant (project \mbox{STRIPES}, Grant agreement ID: \mbox{101165317}, DOI:~10.3030/101165317, PI: D.~Tzionas). 

\vfill

%% file: sec/X_suppl.tex
\renewcommand{\thesection}{S.\arabic{section}}
\renewcommand{\thefigure}{S.\arabic{figure}}
\renewcommand{\thetable}{S.\arabic{table}}
\renewcommand{\theequation}{S.\arabic{equation}}
\makeatletter
\renewcommand{\theHsection}{S.\arabic{section}}
\renewcommand{\theHsubsection}{S.\arabic{section}.\arabic{subsection}}
\renewcommand{\theHfigure}{S.\arabic{figure}}
\renewcommand{\theHtable}{S.\arabic{table}}
\renewcommand{\theHequation}{S.\arabic{equation}}
\makeatother
\setcounter{section}{0}
\setcounter{figure}{0}
\setcounter{table}{0}
\setcounter{equation}{0}

\vspace{-0.5 em}
\section{Implementation Details}

\subsection{MoGe-Based Object Initialization}
\label{subsec:supmat:alignment}

    For each method (D--G) in 
    Tab.~\textcolor{red}{1} (in main), we apply the same initialization, as follows.
    We initialize the object pose $\objR, \objt$ from \samtd~[\textcolor{eccvblue}{71}]. 
    \blue{The rotation estimate is empirically accurate, so} we keep it unchanged. 
    \blue{However, the translation and scale are noisy.} 
    We \blue{refine} these via MoGe-estimated 
    [\textcolor{eccvblue}{76}]
    metric depth. 
    Let the median MoGe depths under 
    \blue{body}
    and object masks 
    [\textcolor{eccvblue}{39}] 
    be $\medianHumDepth$, $\medianObjDepth$, respectively.
    Their ratio $\rho = \medianObjDepth / \medianHumDepth$ is scale-invariant. 
    Anchoring to the metric body depth $\humDepth$ from CameraHMR 
    [\textcolor{eccvblue}{56}] 
    gives the depth-correction factor 
    $\delta = \rho\,\humDepth / t^\objSuper_z$, where $t^\objSuper_z$ is the z-component of $\objt$.
    \blue{We resolve the scale ambiguity by anchoring the object's metric size to the body dimensions, obtaining $\sigma_\mathrm{moge}=\sigma_{smpl} L^o / L^h$, where $\sigma_{smpl}$ is the height of the template \smpl mesh, $L^\objSuper, L^\humSuper$ are bounding-box heights of the object and body, respectively, as extracted from MoGe depth map. 
    Then, we blend $\sigma_\mathrm{moge}$ with \samtd-estimated scale $\sigma_\mathrm{sam3d}$:}
        $\sigma = (\delta\,\sigma_\mathrm{sam3d})^{0.8}\cdot\sigma_\mathrm{moge}^{0.2}$.

\subsection{Evaluation Metrics}
\label{subsec:supmat:metrics}

    \hspace{\parindent}
    For the human \blue{body}, 
    as in 
    [\textcolor{eccvblue}{52}, \textcolor{eccvblue}{77}], 
    we apply Procrustes alignment on the estimated \smplh mesh (to align it to the GT \smplx mesh), via a similarity transform $\{ s, \mathbf{R}, \mathbf{t} \}$ computed on vertices that are common between \smplh and \smplx. 
    For the object, we apply only the rigid transform $\{ \mathbf{R}, \mathbf{t} \}$ (from above) to preserve and evaluate the estimated scale, $s$. 
    After the above, 
    we compute all metrics on $10{,}000$ points uniformly sampled across the aligned surfaces.
    
    \zheading{Chamfer Distance (CD)} 
    A bidirectional surface-to-surface distance (cm): 
    \begin{equation}
        \small
        \mathrm{CD}(\estV, \gtV) =    \frac{1}{|\gtV|}             \sum_{\gtVertex\in \gtV}\min_{\estVertex\in \estV}\|\gtVertex-\estVertex\| + 
                                                    \frac{1}{|\estV|}   \sum_{\estVertex\in \estV}\min_{\gtVertex\in \gtV}\|\gtVertex-\estVertex\|
        \text{,}
    \end{equation}
    capturing the 
    3D geometric error between an estimated and GT mesh with vertex sets $\estV$ and $\gtV$, respectively. 
    Lower is better ($\downarrow$). 
    
    \zheading{Contact F1} 
    This captures the agreement between the estimated ($\estMask$) and ground-truth ($\gtMask$) binary contact masks, $\gtMask \in \{0, 1\}^N$. 
    For a human \blue{body} surface point $\humSurfacePoint \in \nR^3$, binary contact is defined as $\indicatorFunc(\min_{\objSurfacePoint \in \objSurface} \|\humSurfacePoint - \objSurfacePoint\| \leq 5\text{cm})$ by finding the closest point $\objSurfacePoint$ on the  object surface, $\objSurface$, and thresholding the respective distance. Moreover, $\indicatorFunc(\cdot)$ is the indicator function, with $\indicatorFunc(a<b)=1$ iff $a<b$. 
    Let $TP = |\estMask \cap \gtMask|$ be the number of true positive contact points. 
    We compute:
    \begin{equation}
        \small
        P = \nicefrac{TP}{|\estMask|}   \text{,}    \quad 
        R = \nicefrac{TP}{|\gtMask|}    \text{,}    \quad 
        F1 = \nicefrac{2PR}{P+R}        \text{,}
    \end{equation}
    where $P$ is Precision and $R$ is Recall. 
    Higher is better ($\uparrow$).
    
    \zheading{Collision Score} 
    This quantifies the fraction of body vertices that penetrate into the object mesh:
    \begin{equation}
        \text{Collision}(\humVerts, \objverts) = \frac{1}{|\humVerts|} \sum_{\bv^\humSuper \in \humVerts} \indicatorFunc(\bv^\humSuper \blue{\text{~in~}} %
        \objverts)
        \text{,}
    \end{equation}
    where $\humVerts, \objverts$ are the set of body and object vertices, respectively, 
    $\indicatorFunc(\bv^\humSuper \blue{\text{~in~}} %
    \objverts)$ is a binary indicator of whether a body vertex, $\bv^\humSuper$, is inside the object volume $\objverts$, computed via ray-parity testing 
    [\textcolor{eccvblue}{34}]. 
    Lower is better ($\downarrow$).

\section{Qualitative Evaluation \blue{(Extending Sec.~4.3 of Main)}}
\label{sec:rationale}

\input{fig/supmat_failure_cases}
\input{fig/supmat_additional_comparisons_1}
\input{fig/supmat_additional_comparisons_2}

\subsection{Additional Results}
\label{subsec:supmat:additional_results}

    \hspace{\parindent}
    \zheading{Comparisons to \sota} 
    The new extensive comparisons in \cref{fig:supmat_extra_quat_1,fig:supmat_extra_quat_2} \blue{(which extend Fig.~6 of the main)} demonstrate the robustness of \lexisFlowRefinement across diverse, complex interactions \blue{shown in images taken in the wild}. 
    \blue{For more results shown with a rotating viewpoint, see our \video.} 
    
    \zheading{Failure Cases} 
    \blue{As with all methods,} 
    performance can degrade when poor off-the-shelf initialization exceeds the recovery capacity of the Flow, or for highly atypical \blue{interactions} underrepresented in existing datasets 
    [\textcolor{eccvblue}{41}, 
     \textcolor{eccvblue}{81}, 
     \textcolor{eccvblue}{91}, 
     \textcolor{eccvblue}{93}]. 
    In such \blue{Out-Of-Distribution (OOD)} cases, the \lexisDict manifold (and the decoded \interfields) may misguide \lexisFlow, \blue{causing} floating or penetrating artifacts. 

\input{fig/supmat_perceptual}
\subsection{Perceptual Study}
\label{subsec:supmat:perceptual}

To evaluate perceived realism, we conduct a perceptual study.
We randomly sample 60 images from \opentdhoi~[\textcolor{eccvblue}{78}] and reconstruct
each using our \lexisFlowRefinement and the \sota~\hoigaussian~[\textcolor{eccvblue}{78}] method.
We randomize both the presentation order and left/right placement.
Each image is shown to 62 participants, who select the reconstruction that better matches the input image and appears more physically plausible. 
We present 60 samples, of which 4 are catch trials to assess participant reliability (\eg, to detect participants that do not understand the task); this filters out 1 participant, leaving out a total of 61 valid participants. 
\lexisFlowRefinement is preferred in 75.8\% of comparisons over \hoigaussian.
For participant instructions, see \cref{fig:supmat_perceptual_protocol}.

%% file: fig/supmat_failure_cases.tex
\begin{figure}[b!]
    \centering
    \vspace{-0.5 em}
    \includegraphics[width=0.99 \columnwidth]{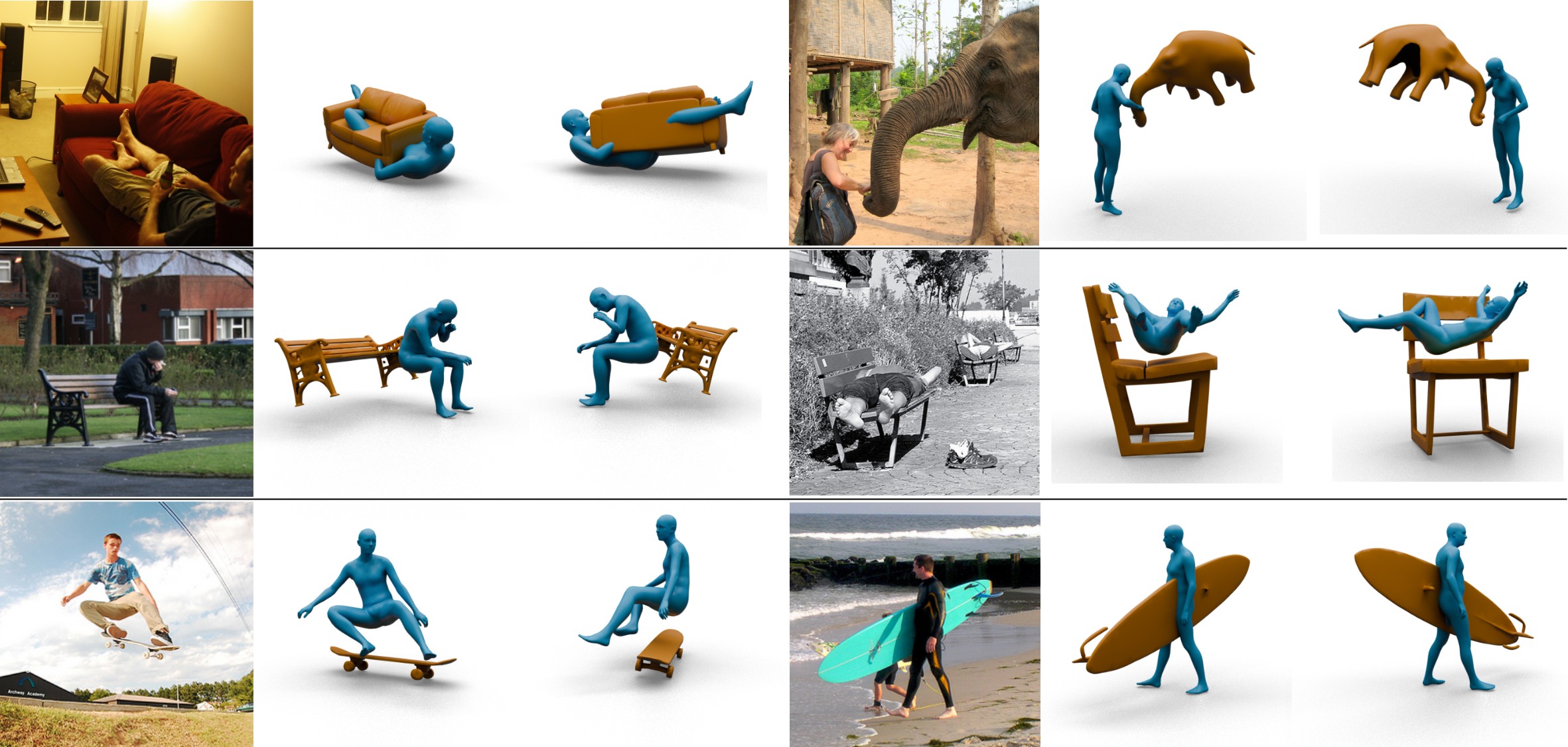}
    \vspace{-1.0 em}
    \caption{
        \qheadingFIGTAB{Failure cases} 
        Errors arise from inaccurate body pose and depth estimation. 
        This can lead to misplaced human–object configurations, \eg, hovering over a bench or skateboard instead of contacting it. 
    }
    \label{fig:supmat_failure_cases}
    \vspace{-1.0 em}
\end{figure}

%% file: fig/supmat_additional_comparisons_1.tex
\newcommand{\supmatCaptionOursVsSota}{
            \qheadingFIGTAB{\lexisFlow vs \sota methods} 
            Across many \inthewild images [\textcolor{eccvblue}{78}] and for 
            a wide variety of objects, our \lexisFlow method 
            estimates more physically-plausible 3D \HOI estimates than existing \sota approaches, 
            capturing better cphysical coupling while reducing floating artifacts and interpenetrations. 
            For more results shown with a rotating viewpoint, see our \video. 
            Here: \faSearch~\textbf{Zoom in} to see 3D details. 
}

\begin{figure}
    \centering
    \begin{overpic}[width=1 \columnwidth, unit=1mm]{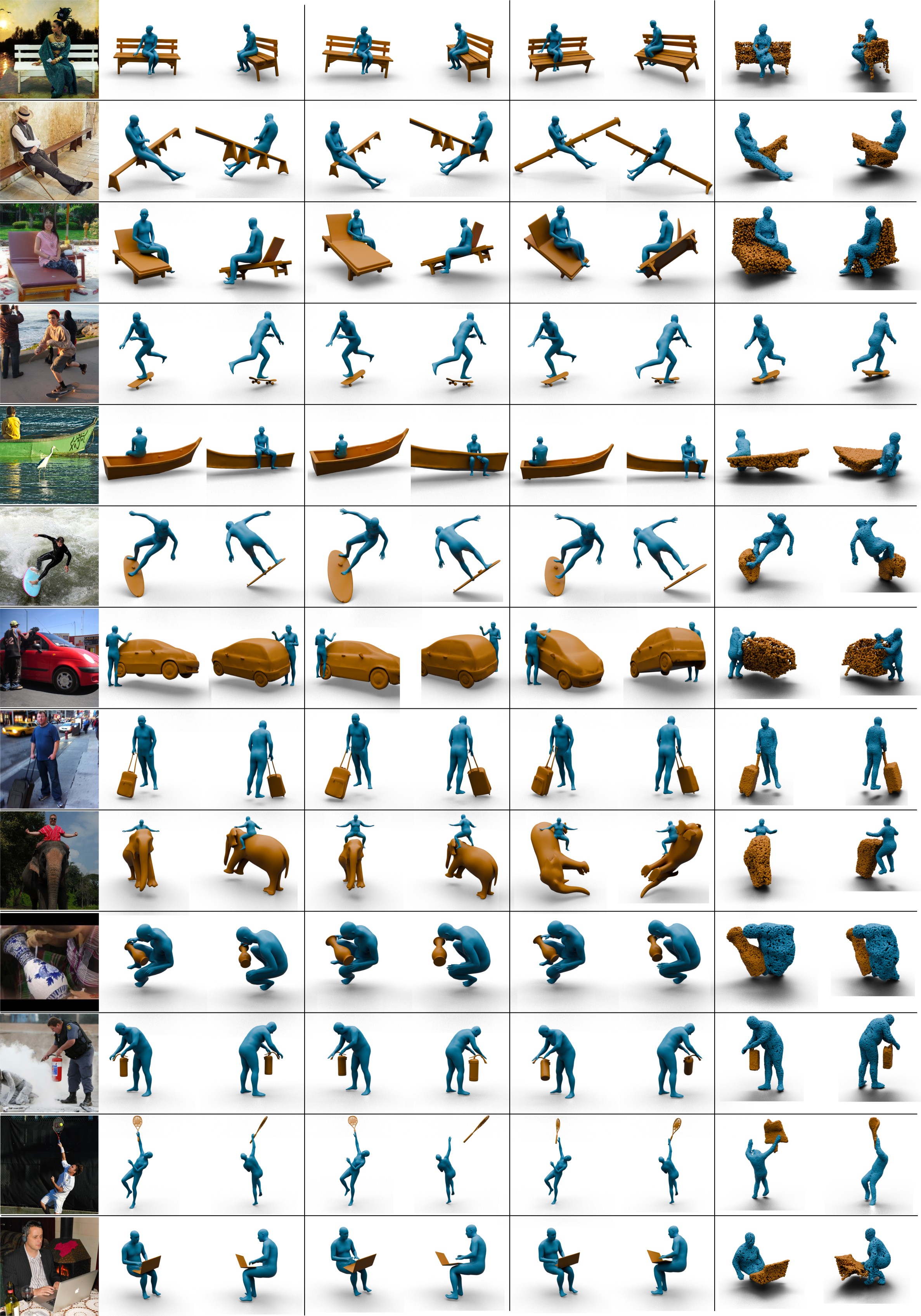}
        \put(2, 101){\tiny Input}
        \put(10, 101){\tiny $\lexisFlow^*$ (ours)}
        \put(26, 101){\tiny \hoigaussian    [\textcolor{eccvblue}{78}]
        }
        \put(42, 101){\tiny \interactvlm    [\textcolor{eccvblue}{16}]
        }
        \put(59, 101){\tiny \hdm            [\textcolor{eccvblue}{81}]
        }
    \end{overpic}
    \vspace{-1.0 em}
    \caption{\supmatCaptionOursVsSota}
    \label{fig:supmat_extra_quat_1}
\end{figure}

%% file: fig/supmat_additional_comparisons_2.tex
\begin{figure}
    \centering
    \begin{overpic}[width=1 \columnwidth, unit=1mm]{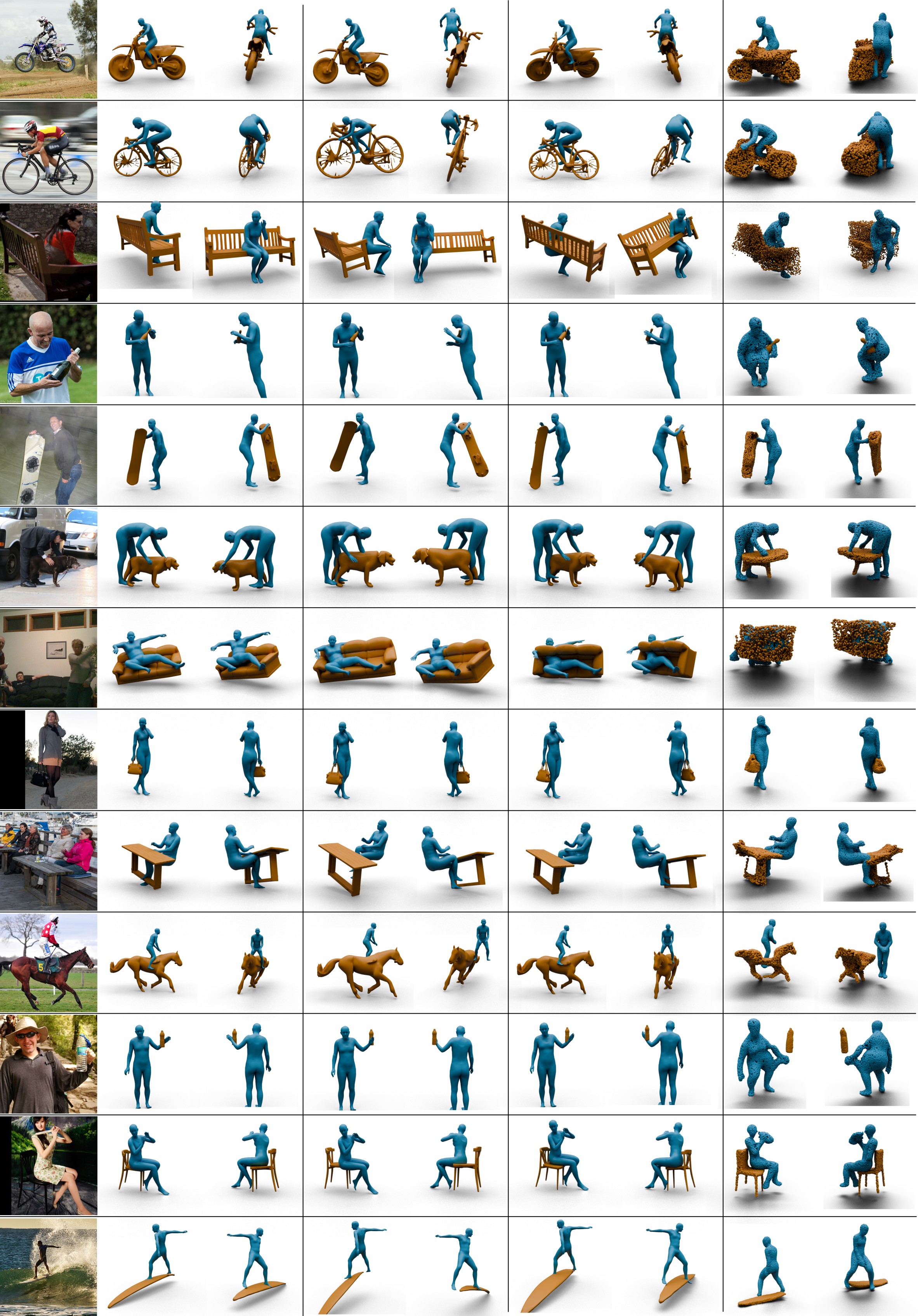}
        \put(2, 101){\tiny Input}
        \put(10, 101){\tiny $\lexisFlow^*$ (ours)}
        \put(26, 101){\tiny \hoigaussian    [\textcolor{eccvblue}{78}]
        }
        \put(42, 101){\tiny \interactvlm    [\textcolor{eccvblue}{16}]
        }
        \put(59, 101){\tiny \hdm            [\textcolor{eccvblue}{81}]
        }
    \end{overpic}
    \vspace{-1.0 em}
    \caption{\supmatCaptionOursVsSota}
    \label{fig:supmat_extra_quat_2}
\end{figure}

%% file: fig/supmat_perceptual.tex
\begin{figure}[t]
    \centering
    \vspace{-0.5 em}     %
    \includegraphics[trim={00mmm 25mm 00mm 70mm},clip, width=0.95 \columnwidth]{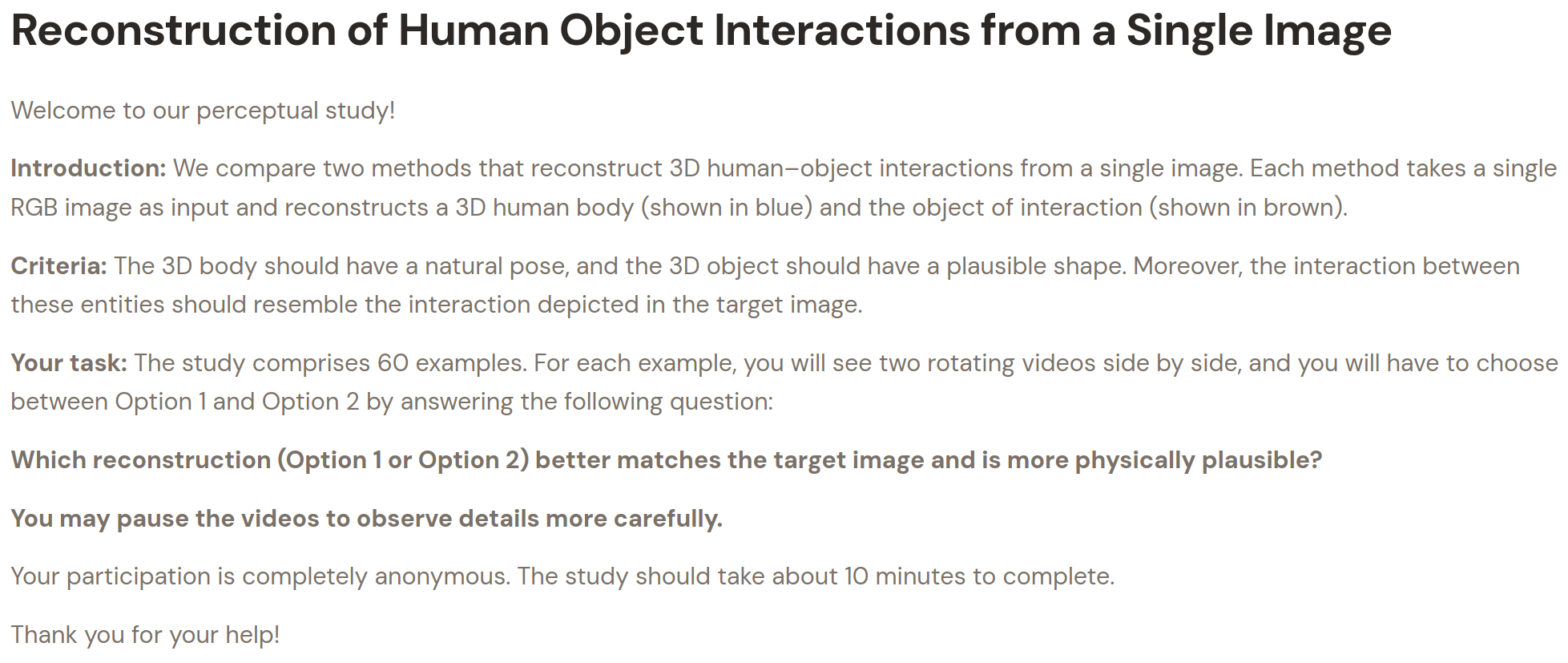}
    \vspace{-1.0 em}
    \caption{
        \qheadingFIGTAB{Perceptual study protocol -- Instructions to participants} 
        We ask 62 participants to view 60 images and respective 3D reconstructions by our \lexisFlowRefinement method and the \sota~\hoigaussian~[\textcolor{eccvblue}{78}] method. 
        Then, for each image they select the reconstruction that better matches the image and appears more physically plausible. 
    }
    \label{fig:supmat_perceptual_protocol}
    \vspace{-1.0 em}
\end{figure}